\documentclass{article}
\usepackage{iclr2027_conference,times}
\iclrfinalcopy
\usepackage[T1]{fontenc}
\usepackage{amsmath,amssymb,amsthm,graphicx,booktabs,tabularx,array,longtable}
\usepackage{hyperref,xurl,placeins,etoolbox}
\hypersetup{hidelinks,pdftitle={LACS: Distribution-Sensitive Label-Agnostic Confidence Selection for Dynamic Data Pruning},pdfauthor={Heejin Choi},pdfsubject={Preprint}}

\newcommand{\LACS}{\textsc{LACS}}
\newcommand{\lacst}{\textsc{LACS-T}}
\newcommand{\ind}{\mathbf{1}}

\newcommand{\sdD}{\mathrm{SD}_{D}}
\newcommand{\sdS}{\mathrm{SD}_{s}}
\newsavebox{\resulttablebox}
\newcommand{\tabinput}[1]{\sbox{\resulttablebox}{\input{tables/#1}}\ifdim\wd\resulttablebox>\linewidth\resizebox{\linewidth}{!}{\usebox{\resulttablebox}}\else\usebox{\resulttablebox}\fi}
\title{LACS: Distribution-Sensitive Label-Agnostic Confidence Selection\\for Dynamic Data Pruning}
\author{Heejin Choi\\
Yonsei University \quad Sooil Development \quad NCreate\\
\texttt{heejin.share@gmail.com}}
\begin{document}
\maketitle
\lhead{Preprint}
\begin{abstract}
Training on large datasets motivates selecting which examples receive expensive training computation. Fixed quotas determine the amount of pruning in advance, while absolute confidence cutoffs depend on the score scale. We study a population-relative alternative that adapts the threshold and retained count together.

We introduce \LACS\ (Label-Agnostic Confidence Selector), a training-time rule based on maximum predictive confidence, requiring neither labels at selection time nor per-example gradients. LACS removes examples at or above a log-mean-exp reference plus a margin. We characterize its sensitivity to score gaps and mass, and derive an exact no-pruning condition. For $0<\rho<1$, the removed fraction is at most $\rho$ and at least one example remains. The unsigned score permits both confidently correct and confidently wrong examples to be removed.

Seven evaluation settings include neural training, language modeling, and boosting. Neural execution studies reduce backward-active rows by 10--35\%; on CIFAR-10, 29\% fewer backward rows accompany accuracy means of 89.77\% versus 89.82\% for full CE. At a matched backward budget and number of optimizer updates, LACS is 1.0 pp above small-batch training and 1.2 pp above random selection over five seeds. Across 14 tabular datasets, quadratic boosting reduces mean test error from 23.36\% to 22.96\%. Its full training replays use about 30\% fewer fitting rows; these counters span a different horizon from validation-selected accuracy. In one RBF-SVM case study, elapsed time falls by 37\% with slightly lower error. These findings separate selective-work reduction from realized acceleration: the current exact-current-score neural implementation uses an additional scoring forward, so its backward sparsity does not yet translate into wall-clock acceleration. This overhead is an execution choice rather than a requirement of the selection rule.

\end{abstract}
\section{Introduction}
\label{sec:intro}
Modern learning increasingly relies on datasets whose scale makes data efficiency a central computational concern \citep{qin2024infobatch,zhou2025seta,raju2021dynamic}. The training value of an example changes as the model learns. Under cross-entropy, a confidently correct prediction has a small derivative with respect to its logits, while a confident mistake can carry a much larger derivative. Omitting examples can therefore change both training cost and the fitted predictor. We ask when a simple confidence rule can reduce selected training work while retaining useful predictive performance.

We study \emph{training-time instance subset selection}: at each selection event, a rule chooses which candidates enter backward propagation or weak-learner fitting. Our primary measurements count participation in these operations. Scoring forwards, optimizer steps, and elapsed time are recorded separately: a reduction in one component is not a reduction in total work by definition. This distinction lets us evaluate a selection policy and its execution overhead without treating them as the same scientific question.

Selection is delicate because removing examples changes the update and subsequent model trajectory. A fixed quota specifies the amount removed before examining the current score gaps. An absolute threshold instead ties the decision to the confidence scale. Population-relative methods offer another design choice: InfoBatch uses a mean-loss reference to define a low-loss pool, then thins that pool stochastically and rescales surviving contributions \citep{qin2024infobatch}. We study a different reference and selection contract, based on unsigned confidence rather than loss.

\begin{figure}[t]
\centering
\begin{minipage}[t]{.49\linewidth}
\centering
\includegraphics[width=\linewidth]{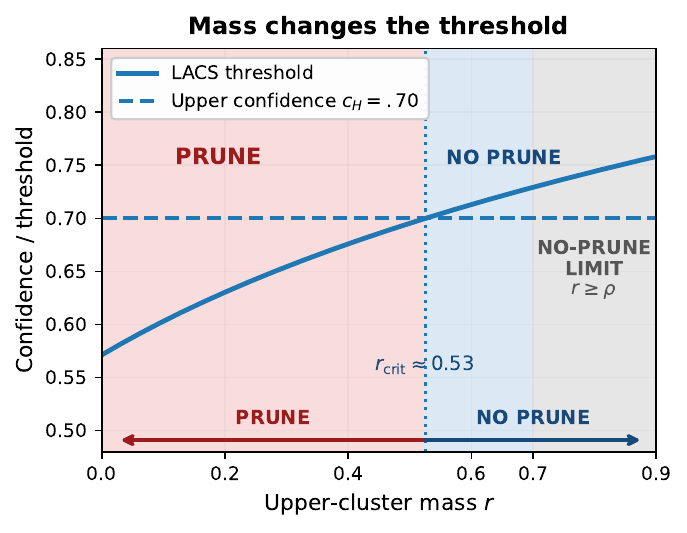}
\end{minipage}\hfill
\begin{minipage}[t]{.49\linewidth}
\centering
\includegraphics[width=\linewidth]{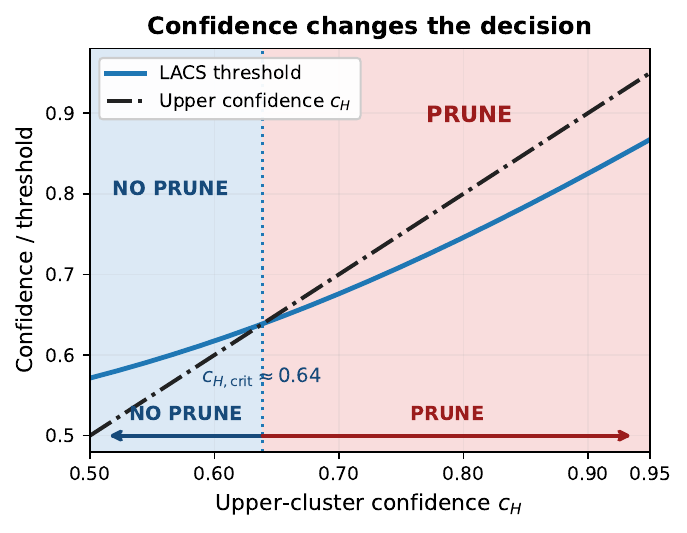}
\end{minipage}
\caption{\textbf{Two one-dimensional slices of the illustrative two-cluster geometry.} \textbf{Left: fixed confidences, varying upper-cluster mass.} With $c_L=.50$, $c_H=.70$, $\tau=.20$, and $\rho=.7$, the adaptive threshold rises with mass and crosses the fixed upper confidence at $r_{\rm crit}\approx.53$; the upper cluster is pruned at or left of the crossing and retained to the right.  \textbf{Right: fixed mass, varying upper-cluster confidence.} With $r=.40$, the threshold and $c_H$ cross at $c_{H,\rm crit}\approx.64$; insufficient separation gives no pruning; at or above the crossing the upper cluster is pruned. Background colors indicate the cluster decision in each slice.  These are illustrative slices, not assumptions made by LACS; $\tau=.20$ is chosen here to keep the transition away from the plot boundary, whereas experiments use the protocol-specific frozen or selected $\tau$.  Appendix~\ref{app:proofs} gives the full $(r,\Delta)$ phase diagram.}
\label{fig:geometry-main}
\end{figure}

\LACS, the \emph{Label-Agnostic Confidence Selector}, compares each candidate's exponentiated maximum confidence with the current population mean of exponentials. Clipping the resulting coefficient at zero produces a sparse support. The equivalent confidence cutoff is a log-mean-exp reference plus a margin; its position and the retained count are outcomes of the current scores. At the same arithmetic mean \emph{confidence}, different score shapes can induce different cutoffs. A maximum-centered mass gives the exact no-pruning condition, and a two-cluster example illustrates how gap and mass determine the decision (Figure~\ref{fig:geometry-main}). These are properties of the proposed cutoff, not a presumption that adaptive counts outperform tuned fixed ones.

Our seven evaluation settings are (i) supervised neural training, (ii) Transformer language modeling, (iii) long-tailed classification, (iv) semi-supervised learning, and boosting with (v) compact nonlinear, (vi) tree, and (vii) kernel-SVM weak learners. The implementations distinguish normalized LACS from the fixed-threshold tree companion \lacst. Boosting is informative because selection precedes every new weak-learner fit. The strongest positive prediction results are in selected fitting; neural comparisons quantify backward sparsity and execution overhead. Additional settings and row-level diagnostics---wrong removals, re-entry, label-flip enrichment, and one-sided interventions---are reported in the supplement.

\paragraph{Contributions.}
(1) A label-agnostic confidence selector with a closed-form population-relative cutoff and an $O(n)$ implementation once confidences are available. (2) An analytical characterization of this cutoff, including the exact no-pruning condition, nonempty-support and cardinality guarantees for $0<\rho<1$, and gap/mass decision boundaries. The standard log-mean-exp facts provide interpretation rather than a separate novelty claim. (3) Controlled empirical evaluations with explicit method variants, uncertainty, and cost horizons: selected fitting yields modest predictive gains in several comparisons, while neural studies distinguish reduced backward participation from full training cost. Extensive additional controls and boundary results appear in the appendix.

\section{LACS: distribution-sensitive instance selection}
\label{sec:method}
\subsection{Confidence, relative scores, and sparse support}
Let $\mathcal B_t=\{x_i\}_{i=1}^{n}$ be the candidate population at selection event $t$, and let $p_{\theta_t}(k\mid x_i)$ be the model's predictive probability.  We use standard maximum confidence
\begin{equation}
 c_{t,i}=\max_k p_{\theta_t}(k\mid x_i),\qquad
 q_{t,i}=\frac{\exp(c_{t,i}/\tau)}{n^{-1}\sum_j\exp(c_{t,j}/\tau)},\qquad
 u_{t,i}=[1-\rho q_{t,i}]_+.
\label{eq:lacs}
\end{equation}
Here $[a]_+=\max(a,0)$, $\tau>0$, and $\rho>0$.  The active set and count are
\begin{equation}
S_t=\{i:u_{t,i}>0\},\qquad K_t=|S_t|.
\label{eq:support}
\end{equation}
Selection is \emph{label-agnostic}: Eq.~\eqref{eq:support} never reads $y_i$ (the downstream learner and the hyperparameter search remain supervised).  Subtracting $\max_i c_{t,i}$ inside both exponentials gives a numerically stable implementation.

Two downstream contracts are used. \emph{Hard support} passes $S_t$ to the fitter with its ordinary fitting weights. \emph{Continuous weighting} uses $w_{t,i}=u_{t,i}/\bar u_t$, where $\bar u_t=n^{-1}\sum_j u_{t,j}$; neural experiments detach these coefficients. The analysis below uses $0<\rho<1$, which guarantees $\bar u_t>0$. Historical searches over a wider domain are identified in Appendix~\ref{app:protocols}.

For detached coefficients and sample-independent computations, exact physical omission targets the dense weighted update
\begin{equation}
\nabla_\theta\left[\frac1n\sum_i \operatorname{stopgrad}(w_i)\ell_i(\theta)\right]
=\frac1n\sum_{i\in S}w_i\nabla_\theta\ell_i(\theta).
\label{eq:exact}
\end{equation}
Here $\operatorname{stopgrad}(w_i)$ holds the current coefficients fixed. A zero-weight term contributes no gradient; omitting its computation preserves the frozen weighted update when retained activations and Jacobians, the original denominator, and stochastic realizations stay unchanged. Training-mode BatchNorm couples examples through shared batch statistics, so physical removal can change retained gradients \citep{ioffe2015batchnorm}. Appendix~\ref{app:proofs} details these conditions and the numerical checks. This identity concerns execution of the LACS-weighted surrogate, separately from its predictive quality relative to unweighted CE.

Given scalar confidences, selection needs $O(n)$ arithmetic and no sorting or subset solver. Computing these confidences can require additional model execution. Equal support also need not yield equal updates: survivor weights, normalization, and boosting coefficients are part of the downstream contract.

\subsection{A threshold determined by the current distribution}
For $0<\rho<1$, write $\kappa=1/\rho>1$.  An example is removed exactly when
\begin{equation}
 c_i\ge T(c),\qquad
 T(c)=\tau\log\!\left(\frac1n\sum_j e^{c_j/\tau}\right)+\tau\log\kappa.
\label{eq:cutoff}
\end{equation}
Thus LACS is a thresholding rule, but not an absolute-confidence rule: its threshold moves with the whole confidence population.

\paragraph{Why log-mean-exp is more than a mean.}
Set $\beta=1/\tau$ and write
\[
\operatorname{LME}_{\beta}(c)=\frac1\beta\log\!\left(\frac1n\sum_i e^{\beta c_i}\right),
\qquad T(c)=\operatorname{LME}_{\beta}(c)+\frac{\log\kappa}{\beta}.
\]
\paragraph{Fact: same-mean dispersion sensitivity.}
For every $\beta>0$, $\operatorname{LME}_{\beta}$ is symmetric and convex, hence Schur-convex: among score populations with the same mean, a more dispersed vector in the majorization order cannot lower the reference.  Around $\beta=0$,
\begin{equation}
\operatorname{LME}_{\beta}(c)
=\mu+\frac{\beta}{2}\sigma^2+\frac{\beta^2}{6}\kappa_3+O(\beta^3),
\label{eq:cumulant}
\end{equation}
where $\kappa_3$ is the empirical third cumulant.
The reference also satisfies $\mu\le\operatorname{LME}_{\beta}(c)\le c_{\max}$, tending to $\mu$ as $\beta\to0$ and to $c_{\max}$ as $\beta\to\infty$.  A boundary using only the arithmetic mean of the \emph{same score} is identical for same-mean populations; the LACS boundary separates them through spread and upper-tail shape: concentration near the mean pulls the reference toward $\mu$, whereas dispersion and, locally, positive upper-tail skew raise it.  These reference limits do not apply to the full cutoff: at fixed $\kappa>1$, the added $\log\kappa/\beta$ diverges as $\beta\to0$. A higher reference also need not imply fewer removals, because the scores can move with the boundary.  Appendix~\ref{app:proofs} gives the proof and same-mean examples.

Let $c_{\max}=\max_i c_i$ and define the maximum-centered mass
\begin{equation}
M_\tau(c)=\sum_i e^{(c_i-c_{\max})/\tau}.
\label{eq:massdef}
\end{equation}
Then
\begin{equation}
\boxed{D=\varnothing\ \Longleftrightarrow\ M_\tau(c)>\rho n},
\qquad |D|/n\le\rho.
\label{eq:mass}
\end{equation}
Each term in $M_\tau$ lies in $(0,1]$: examples near the maximum count almost fully and distant examples count less.  Hence a homogeneous population is never pruned, whether its common confidence is low or high.  The criterion is about how much mass lies near the current upper end, not about total confidence ($\sum_iq_i=n$ is fixed).  Proofs and boundary cases appear in Appendix~\ref{app:proofs}.

\subsection{Two-cluster illustration: mass and confidence play different roles}
The rule in Eq.~\eqref{eq:mass} applies to arbitrary score distributions.  To make its geometry interpretable, consider an illustrative distribution with lower-cluster confidence $c_L=a$ and upper-cluster confidence $c_H=a+\Delta$, where the upper cluster has mass $r$.  The normalized near-upper mass is
\begin{equation}
 m_{\mathrm{eff}}(r,\Delta)=r+(1-r)e^{-\Delta/\tau}=M_\tau(c)/n.
\label{eq:effective-mass}
\end{equation}
The complete tied upper cluster is pruned iff $m_{\mathrm{eff}}\le\rho$, equivalently
\begin{equation}
 r<\rho,\qquad
 \Delta\ge\Delta_{\mathrm{crit}}(r)=\tau\log\!\frac{1-r}{\rho-r}.
\label{eq:twolevel-main}
\end{equation}
The raw upper mass $r$ is therefore only part of the criterion.  A small gap makes lower-cluster points contribute to the effective mass; a large gap discounts them.  Increasing upper-cluster mass raises the threshold. At fixed $r<\rho$, a larger gap moves $c_H$ faster than the threshold and can trigger removal; bounded confidence may limit the reachable gap.

\paragraph{Relationship to quantiles.}
At the realized $K_t$, hard LACS support retains the $K_t$ lowest-confidence examples, so the ranking is that of a confidence quantile with the same count.  The difference is that a fixed quantile chooses the count first, whereas LACS makes the quantile level an outcome of the score distribution: LACS is a \emph{gap- and mass-sensitive quantile level}, not a new ranking.  Common additive shifts leave the rule unchanged; arbitrary monotone recalibration need not, an invariance that fixed quantiles do have.

\subsection{Applications and variants}
A generic interface is $S_t=\operatorname{Select}(\mathcal B_t,p_{\theta_t})$, followed by an application-specific update on $S_t$.  Neural selective execution applies continuous coefficients and, when physically skipping rows, preserves the declared original denominator.  Repeated fitting passes selected rows to a tree, SVM, or compact nonlinear weak learner.  Boosting is a particularly revealing instance because selection is repeated before every weak learner is constructed.

The tree experiments use a companion fixed unsigned-margin rule $S_t^T=\{i:|F_t(x_i)|\le\vartheta\}$, denoted \lacst.  Because $\sigma(|F_t|)$ is the confidence of a binary logistic prediction, LACS-T is an absolute-confidence variant, not the distribution-sensitive rule of Eq.~\eqref{eq:lacs}; its results support unsigned-confidence selection in general but do not test the normalized cutoff, and we label them accordingly.  Complete score, weighting, parameter, and denominator contracts are listed in Appendix~\ref{app:protocols}.

\section{Interpreting the unsigned confidence rule}
\label{sec:analysis}
The same unsigned score can have different update-level meanings. For binary cross-entropy, write $m_i=y_iF_i$, $y_i\in\{-1,+1\}$, and $c_i=\sigma(|F_i|)$. Then
\begin{equation}
\left|\frac{\partial\ell_i}{\partial F_i}\right|
=\begin{cases}1-c_i,&m_i>0,\\c_i,&m_i<0.\end{cases}
\label{eq:gradient_cases}
\end{equation}
A confidently correct row has a small output derivative; a confidently wrong row retains a large one. Parameter-gradient size additionally depends on the model Jacobian. For hard support with the original denominator, let $G=n^{-1}\sum_i g_i$, $G_S=n^{-1}\sum_{i\in S}g_i$, and $\|\nabla_\theta F_i\|\le L_i$. Splitting removed rows into correct and wrong sets $D_+$ and $D_-$ yields
\[
\|G-G_S\|\le\frac1n\left[\sum_{i\in D_+}L_i(1-c_i)+\sum_{i\in D_-}L_i c_i\right].
\]
This conditional bound connects correct-tail removal to gradient fidelity. Wrong-tail removal can instead suppress a large original contribution; continuous survivor reweighting adds another term (Appendix~\ref{app:gradient}). The bound does not require either operation to improve prediction.

At the inspected neural checkpoints, removed rows are all correct and the original-denominator gradient residual is 0.4--3.6\%. In tree trajectories, some removed rows are wrong and later re-enter. One-sided interventions have severe correct-only failure cases on two datasets, while aggregate unified-versus-one-sided contrasts remain unresolved. These observations motivate examining the composition of the retained fit; they do not establish that wrong-tail trimming causes the observed accuracy gains. Appendix~\ref{app:onesided} reports the complete interventions, row histories, and label-flip diagnostics.

\paragraph{The no-pruning condition appears in actual trajectories.}
The exact mass criterion is not only a two-cluster illustration. On CIFAR-10, we logged 480 real batches from epochs 1, 10, 25, and 50 across three seeds. The condition $M_\tau(c)/(\rho n)>1$ holds on 108 batches, and exactly those 108 batches prune no row (agreement 1.000); all occur at epoch 50. The epoch-mean retained fraction falls from .886 at epoch 1 to a minimum near .573 around epochs 38--40, then rises to .984 at epoch 50 as the fraction of no-prune batches reaches .934. The same check is exact on logged boosting trajectories (171/171 no-prune rounds for \texttt{pol} and 1/1 for \texttt{electricity}). This late conservative phase is a direct consequence of the population geometry, not a claim that stopping pruning late is universally optimal. Appendix~\ref{app:trajectory} gives the trajectories and real same-mean examples.

RobustBoost and ArchBoost address related influence-control questions using label-signed margins and modified potentials \citep{freund2009robust,li2018archboost}. Label-agnostic trimming itself also has a direct boosting precedent: LogitBoost's weights $p_i(1-p_i)$ are small at either prediction extreme \citep{friedman2000additive}. LACS studies a different population-relative cutoff; Section~\ref{sec:related} distinguishes its criterion from those methods.

\section{Experiments: sparse work, prediction, and realized speed}
\label{sec:results}
\subsection{Overview}
The seven settings of Section~\ref{sec:intro} use the declared confidence-selection contracts and test different learning operations; tree results use the companion LACS-T.  The main text reports the four that most directly answer the paper's two questions, how much expensive work is avoided and what happens to prediction: compact repeated fitting (Section~\ref{sec:compact}), tree boosting (Section~\ref{sec:trees}), a direct neural comparison with dynamic-pruning baselines (Section~\ref{sec:direct}), and the systems accounting that separates work from time (Section~\ref{sec:work}).  Table~\ref{tab:overview} previews them.  The long-tailed and semi-supervised settings are boundary studies rather than headline wins and are reported in the supplement, together with seed-level values, additional baselines, null results, protocol details, and the mechanism diagnostics of Section~\ref{sec:analysis}.

\begin{table}[t]
\caption{\textbf{Four representative results from seven evaluated settings.} Work counters are reported separately from wall time.  Multi-dataset SD is across dataset means; single-run cases report no SD.  $\dagger$: corrected $p<.05$ in the stated comparison family.}
\label{tab:overview}
\centering\small
\begin{tabularx}{\linewidth}{@{}p{.20\linewidth}p{.23\linewidth}p{.24\linewidth}X@{}}\toprule
Setting & Work retained / saved & Reference $\rightarrow$ selected method & Statistical scope \\\midrule
Quadratic boosting, R2 & replay rows $\approx70\%$ of full & error $23.36\pm11.46\rightarrow22.96\pm11.27$ & $\Delta=-.401$ pp, CI $[-.721,-.081]$, post-hoc Holm $p\approx.008^{\dagger}$ \\
Tree boosting (LACS-T), depth 2 & same per-round $K_t$ as GOSS & error $20.61\pm11.00\rightarrow20.27\pm11.01$ & $\Delta=-.346$ pp, CI $[-.604,-.088]$, new-block Holm $p=.0147^{\dagger}$; LACS-T variant \\
RBF-SVM boosting & fit rows $3.000\rightarrow2.121$ M & error $24.936\rightarrow24.702$; wall $3400\rightarrow2136$ s & one common-endpoint replay; no multi-seed significance claim \\
CIFAR-10 neural & backward rows $2.496\rightarrow1.776$ M (28.9\% fewer) & accuracy $89.82\pm.20\rightarrow89.77\pm.51$ & three seeds; CI crosses zero; current two-pass wall is slower \\
\bottomrule
\end{tabularx}
\end{table}

\paragraph{Uncertainty conventions.}
For a single dataset, $\pm$ is the sample SD over the stated seeds.  For multi-dataset aggregates, $\pm$ is the SD across dataset means after averaging the prescribed resplits; it measures task heterogeneity, not seed noise.  Paired intervals are nominal 95\% $t$ intervals for the mean difference with the dataset (or seed) as the unit.  Exact sign-flip tests and Holm families are used only where they were recorded or explicitly recomputed; Appendix~\ref{app:stats} gives the complete procedures.  An interval that crosses zero is reported as unresolved, not as evidence of equivalence.

\subsection{Selected fitting can improve prediction}
\label{sec:compact}
The clearest result for the normalized rule comes from \emph{compact repeated fitting}: boosting with small nonlinear weak learners on 14 tabular datasets with three resplits.  A rank-$R$ quadratic learner has raw predictor $b+w^\top x+\sum_{j=1}^{R}a_j(u_j^\top x)^2$ followed by a bounded output (denoted R1, R2, R4); one-hidden-layer MLP learners (H1, H2, H4) and all six families are reported in Appendix~\ref{app:compact}.  Before each round LACS selects the support from the current ensemble confidence; ordinary logistic fitting weights are kept on that support, and the boosting coefficient is estimated on it.  Full-data and LACS procedures are tuned independently on validation data, so this tests the complete selected-training procedure rather than a membership-only intervention.

\begin{table}[t]
\caption{\textbf{Quadratic repeated fitting across 14 datasets.} Error is mean $\pm\sdD$.  $\dagger$ marks $p<.05$ under the reported post-hoc Holm correction of the six full-data comparisons; intervals are nominal, not simultaneous.  The supplement reports all random-support controls and per-dataset results.}
\label{tab:compact}
\centering\small\begin{tabular}{@{}lrrrrr@{}}\toprule
Weak learner & Full error & LACS error & $\Delta_{\rm err}$ & 95\% $t$ CI & $p_{\rm H,full}$ \\\midrule
Quad R1 & $23.64\pm11.30$ & $23.34\pm11.19$ & $-0.294$ & $[-0.623,+0.035]$ & $0.281$ \\
Quad R2 & $23.36\pm11.46$ & $22.96\pm11.27$ & $-0.401^{\dagger}$ & $[-0.721,-0.081]$ & $0.008$ \\
Quad R4 & $23.19\pm11.38$ & $22.81\pm11.23$ & $-0.374^{\dagger}$ & $[-0.664,-0.084]$ & $0.014$ \\
\bottomrule
\end{tabular}

\end{table}

Quadratic R2 is the strongest full-data result (Table~\ref{tab:compact}): mean error falls from $23.36$ to $22.96\%$, a paired difference of $-.401$ pp with nominal CI $[-.721,-.081]$.  R4 also improves under the same post-hoc correction; the MLP versus full-data comparisons are unresolved.  The per-round count control is an exact-count random support at the LACS configuration, which removes the same number of rows per round: R2 wins on 13/14 datasets and passes the original six-learner Holm family ($p=.0012$), and H2, R1, and R4 pass it as well.  This supports the complete LACS selected-fitting procedure against that random control. It does not isolate membership alone: the stage coefficient is estimated on each support, and each arm chooses its own validation stopping round.

A \emph{replay} is the full recorded training run; its \emph{selected prefix} ends at the checkpoint chosen on validation data. These are different accounting horizons.  Over complete replays, R2 fits about .70 of the full-training rows.  At the validation-selected predictor, the pooled selected-prefix ratio is .641 but the mean of per-run ratios is .936, because stopping rounds and dataset sizes differ.  Selected-prefix wall time was not recorded, so we make no time-to-equal-quality claim; Appendix~\ref{app:compact} reports both horizons for all six learner families.

\subsection{Boosting as repeated subset selection}
\label{sec:trees}
Boosting exposes selection repeatedly because a new learner is fit after every selection event.  The tree campaign compares the fixed-threshold companion \lacst\ (Section~\ref{sec:method}) with budget-matched GOSS-style selection and full XGBoost in one harness \citep{chen2016xgboost,ke2017lightgbm}; these results evaluate the fixed-threshold companion, not the normalized cutoff. The partner-config full-data arm is distinct from an independently tuned full-data algorithm.

At depth 2, budget-matched GOSS has mean error $20.61\pm11.00\%$ and LACS-T $20.27\pm11.01\%$, a paired change of $-.346$ pp with nominal CI $[-.604,-.088]$; on the newly added blocks the comparison survives the prespecified three-depth Holm correction ($p=.0147$).  An independently tuned GOSS arm gives a difference of $-.386$ pp (nominal sign-flip $p=.031$) and is reported as a secondary comparison.  Depths 1 and 3, the full-data arms, a ten-resplit depth-one study, and per-dataset results are in Appendix~\ref{app:trees}.  The point is not that LACS-T dominates tree sampling, but that repeated confidence selection can improve prediction at a controlled fitting budget.

\subsection{Direct comparison with dynamic pruning baselines}
\label{sec:direct}
The CIFAR-10 study (ResNet-32 with GroupNorm, 50 epochs, three seeds) compares LACS with recent dynamic-pruning methods under one training recipe.  Table~\ref{tab:direct-neural} reports prediction, backward participation, scoring forwards, and optimizer steps. The complete nine-arm results, paired intervals, runtime repeats, and implementation checks are in Appendix~\ref{app:infobatch}.

\begin{table}[t]
\caption{\textbf{Direct neural pruning comparison, three seeds.}  Extra forward rows and steps expose costs not captured by matching backward rows; paired intervals and same-host runtime details are in Appendix~\ref{app:infobatch}.  V is validation-selected, B approximately matches LACS's backward count, and N is a near-budget RePB point.}
\label{tab:direct-neural}
\centering\small
\begingroup\setlength{\tabcolsep}{4pt}
\begin{tabular}{@{}lrrrr@{}}\toprule
Method & Test acc. (\%) $\uparrow$ & Bwd. (M) & Extra fwd. (M) & Steps \\\midrule
Full CE & $89.82\pm0.20$ & 2.496 & 0.000 & 19,500 \\
LACS & $89.77\pm0.51$ & \textbf{1.776} & 2.496 & 19,500 \\
InfoBatch (V) & $89.75\pm0.21$ & 2.034 & 0.000 & 15,893 \\
InfoBatch (B) & $89.06\pm0.33$ & 1.776 & 0.000 & 13,876 \\
RePB (N) & $89.46\pm0.05$ & 1.827 & 0.000 & 14,274 \\
Selective-Backprop & $89.55\pm0.14$ & 1.795 & 2.446 & 14,047 \\\bottomrule
\end{tabular}\endgroup
\end{table}

LACS removes 29\% of backward rows relative to full CE with an accuracy mean $.05$ pp lower. Its mean is $.021$ pp above validation-selected InfoBatch and $.708$ pp above backward-matched InfoBatch-B. The latter difference is positive in all three seeds but unresolved ($p_t=.180$); neither superiority nor equivalence is established. Importantly, LACS takes 19,500 optimizer steps versus 13,876 for InfoBatch-B, with smaller effective batches and an extra scoring forward. Matching backward counts therefore does not isolate selection from update count. The experiment establishes sparse backward participation and close observed accuracy means, not a total-compute advantage.

\paragraph{Selection, not update count.}
A separate five-seed control on the same recipe (Appendix~\ref{app:controls}) removes this confound. Small-batch CE with batch 91 takes 19,764 steps over about 1.80\,M examples, close to LACS's 19,500 steps and 1.78\,M backward rows; a random-drop arm keeps a uniformly random subset of the same size as the LACS support in every batch. LACS is $+1.005$ pp above small-batch CE (95\% CI $[+.572,+1.438]$) and $+1.185$ pp above random drop ($[+.665,+1.705]$), positive in all five seeds, while small-batch CE is itself $.713$ pp below full CE ($[-1.002,-.423]$). Against full CE in this panel LACS is $+.292$ pp ($[-.404,+.989]$). At equal backward budget and update count, the accuracy therefore tracks which examples are kept rather than the smaller effective batch.

\paragraph{CIFAR-100.}
On CIFAR-100 (ResNet-32 with GroupNorm, three seeds; Appendix~\ref{app:controls}), LACS removes 13\% of backward rows at $62.24\pm0.46\%$ versus $62.41\pm0.31\%$ for full CE. Among the budget-matched arms it has the highest mean and is $+.96$ pp above RS2 ($[+.46,+1.46]$); its differences to work-matched InfoBatch ($+.47$, $[-1.13,+2.07]$), near-budget RePB ($+.99$) and Selective-Backprop ($+.48$) are positive but unresolved at three seeds. LACS is slower than InfoBatch on this host (264 versus 190 s).

\paragraph{Which removals carry the accuracy?}
A label-aware diagnostic, C-LACS, removes only the currently correct members of the LACS set. On one host with backward fractions within 1.5\% (Appendix~\ref{app:controls}), C-LACS reaches $90.15\pm.28\%$, LACS $89.74\pm.46\%$, and work-matched InfoBatch $89.25\pm.23\%$; C-LACS exceeds InfoBatch by $+.896$ pp ($[+.119,+1.672]$, positive in all three seeds), while LACS minus InfoBatch ($+.487$, $[-1.222,+2.197]$) is unresolved. Confidently wrong rows are about 1.8\% of the LACS removal set here, so the two supports nearly coincide. C-LACS reads labels and is not the proposed method; the result indicates that deterministic removal of confidently correct examples can match or exceed stochastic low-loss thinning at an equal backward budget, and InfoBatch remains faster on this host (139 versus 221--226 s).

\subsection{Work reduction versus wall-clock speed}
\label{sec:work}
Executed work and elapsed time can move in opposite directions.  If a dense step costs a forward $F$ plus a backward $B$, and an exact-current-batch two-pass implementation first scores the full batch and then trains a retained fraction $r$, then
\begin{equation}
C_{\rm full}=F+B,\qquad C_{\rm 2pass}\approx F+r(F+B)+H,
\label{eq:cost-model}
\end{equation}
where $H$ collects selection and packing overhead.  Under the illustrative dense-layer rule of thumb $B\approx2F$ and $H=0$, the two-pass path saves arithmetic only when $r<2/3$; the CIFAR-10 run retains $r\approx.71$, so this model predicts no saving there. Applying $B=2F$ to the recorded forward and backward row counts gives about 4.5\% more arithmetic than full CE, not a measured FLOP count. This is consistent with the slower runtime below.  Exact ratios depend on architecture and implementation, which is why executed rows and measured wall time are reported separately.  Table~\ref{tab:worktime} shows the two systems regimes we observed.

\begin{table}[t]
\caption{\textbf{Same rule, different systems outcomes.}  Neural LACS removes backward-active examples but the current exact-current-batch implementation adds a full scoring forward; RBF-SVM fitting is expensive enough that fewer fit rows translate directly into elapsed-time savings.}
\label{tab:worktime}
\centering\small
\resizebox{\linewidth}{!}{%
\begin{tabular}{@{}lrrr@{}}\toprule
Setting & Task metric: full $\rightarrow$ LACS & Expensive work: full $\rightarrow$ LACS & Wall: full $\rightarrow$ LACS \\\midrule
CIFAR-10 / ResNet-32 GN & acc. $89.82\pm.20\rightarrow89.77\pm.51$ & bwd. $2.496\rightarrow1.776$ M & $312.9\rightarrow396.8$ s \\
Credit / RBF-SVM boosting & err. $24.936\rightarrow24.702$ & fit rows $3.000\rightarrow2.121$ M & $3400\rightarrow2136$ s \\
\bottomrule
\end{tabular}}
\end{table}

\paragraph{Why the neural implementation uses two forwards.}
To decide from the \emph{current} batch confidence and physically omit backward rows, our implementation runs a no-gradient scoring forward on the candidate batch, forms the support, then runs a gradient-enabled forward/backward on the retained examples only; masking losses on a dense autograd graph would not remove the already-built dense backward kernels.  The second forward is an execution choice, not a requirement of the rule: cached confidences can avoid a fresh scoring pass, while fused selective backward requires different execution support. Such alternatives change the score freshness or implementation and need their own evaluation.  In the measured CIFAR-10 panel this overhead dominates, and LACS is slower than full CE ($396.8$ versus $312.9$ s) and than same-host InfoBatch (Appendix~\ref{app:infobatch}).

\paragraph{A measured speedup without systems optimization.}
Repeated RBF-SVM fitting is expensive enough that the support reduction dominates selection overhead.  Over a fixed 300-round credit replay, LACS fits 29.3\% fewer rows, reduces summed support-vector incidences by 32.7\%, and reduces end-to-end time by 37.2\% while slightly lowering final error.  This is one common-endpoint case study, not a universal speed ratio.  A count-matched random support is faster still but has higher error, so speed and predictive quality remain distinct axes.  A replication over eight dataset/split cells (Appendix~\ref{app:svmrep}) is consistent where all arms run (nearly) the same number of rounds: in four such cells LACS reduces wall time by 30--71\% relative to full fitting. The 89--94\% savings in three other cells mainly reflect earlier stopping, and LACS shows no consistent speed advantage over a same-size random support.

\subsection{Breadth and boundaries}
The remaining settings test whether the rule transfers beyond the headline cases.  In a neural execution matrix over eight dataset/architecture/schedule configurations, exact zeros remove 9.9--35.1\% of backward rows while reproducing the dense LACS-weighted same-state gradient to fp64 precision for GroupNorm/LayerNorm architectures; BatchNorm is the deliberate counterexample (Appendix~\ref{app:execution}).  In Transformer language-model training (800 steps, three seeds), selective execution removes 11.5\% of backward sequences at a validation NLL of $1.569$ versus $1.567$ for dense LACS and $1.520$ for CE, and remains 4.5\% slower than the optimized dense implementation.  Long-tailed combinations, the semi-supervised development study, historical InfoBatch variants, and label-noise and one-sided controls are reported in Appendices~\ref{app:execution}--\ref{app:longtail}.  The matrix establishes conditional execution fidelity; the LM run quantifies both sequence savings and a quality cost relative to CE. Neither is pooled with the CIFAR-10 baseline comparison as a performance-preservation result.

\section{Related dynamic pruning and scope}
\label{sec:related}
\paragraph{Dynamic example selection.}
The direct neural comparisons include Selective-Backprop, RS2, RCAP, InfoBatch, and RePB \citep{jiang2019selective,okanovic2024rs2,hassan2025rcap,qin2024infobatch,zhou2026repb}. Raju et al., SeTa, BLS, and OrderDP provide further dynamic-pruning context \citep{raju2021dynamic,zhou2025seta,zhou2026bls,jin2026orderdp}; DropSample, calibration-oriented pruning, and Online Batch Selection establish earlier confidence- or loss-based reselection \citep{yang2016dropsample,patra2023dynamic,loshchilov2015online}. InfoBatch defines eligibility using a mean of cached losses, thins the eligible pool stochastically, rescales survivors, and returns to full data in its final phase. SeTa uses loss clusters and a scheduled window. Both already depend on score distributions. LACS instead studies a maximum-confidence exponential reference, deterministic clipping, and the resulting gap/mass geometry. The same-mean comparison concerns arithmetic mean \emph{confidence}, not every possible transformation of the scores: under CE, mean loss corresponds to a geometric mean of true-label probabilities (Appendix~\ref{app:related_extended}). Empirical comparisons are in Appendix~\ref{app:infobatch}.

\paragraph{Boosting and influence control.}
Weight trimming and influence trimming are direct precedents for omitting examples during weak-learner fitting \citep{friedman2000additive,friedman2001greedy}. In particular, LogitBoost weights $p_i(1-p_i)$ do \emph{not} require the class label at selection time: they are small for either confidently correct or confidently wrong predictions. For binary confidence $c_i=\max(p_i,1-p_i)$, the weight equals $c_i(1-c_i)$, decreasing on $[1/2,1]$. Thus high-confidence removal and dynamic re-entry are established ideas. LogitBoost trims a prescribed fraction of cumulative working-weight mass; LACS thresholds each exponential-confidence contribution relative to its population average. The cutoff policy, rather than an unprecedented unsigned ranking, is the object studied here.

RHO-LOSS selects using reducible holdout loss \citep{mindermann2022rho}; OHEM, Focal Loss, and Selective-Backprop emphasize hard examples \citep{shrivastava2016ohem,lin2017focal,jiang2019selective}. RobustBoost and ArchBoost limit extreme-error influence through modified potentials or losses \citep{freund2009robust,li2018archboost}. These methods motivate distinct comparisons rather than interchangeable guarantees. Appendix~\ref{app:related_extended} also covers static pruning, gradient matching, and importance sampling.

\paragraph{Contribution in this context.}
We characterize and evaluate a particular normalized confidence cutoff, including its exact no-pruning condition and adaptive retained count. The analysis describes how the rule responds to a distribution; controlled downstream comparisons determine whether that response is useful. At an identical realized count, the hard support agrees with confidence-quantile selection.

\paragraph{Limitations.}
The neural head-to-head studies have three seeds on CIFAR-10 and CIFAR-100, and the two-pass implementation is slower than full CE and InfoBatch. The tree comparison uses LACS-T; the compact-learner search also includes $\rho\ge1$, outside the nonempty-support guarantee. Its performance is not evidence for a guarantee restricted to $0<\rho<1$. The RBF-SVM timing evidence rests on four cells with (nearly) equal round counts. Current direct cutoff controls include a setting where tuned fixed selection is better. Broader adaptive-count comparisons and score-refresh execution are required to strengthen those conclusions; the update-matched control covers one CIFAR-10 recipe. Long-tail and SSL studies provide boundaries rather than positive evidence across all seven settings.

\section{Conclusion}
LACS constructs training supports with a confidence cutoff whose position and retained count respond jointly to the current score distribution. The analysis characterizes this response, while experiments identify modest predictive gains in selected-fitting settings and substantial backward sparsity in neural execution. The two-pass neural implementation is slower than dense training, whereas a fitting-heavy SVM case shows that selective work can translate into elapsed-time savings.

\label{main:end}
\clearpage
\section*{AI Use Statement}
In this work, we used generative AI tools for several research and writing tasks. The tools assisted with developing and refining conceptual and theoretical framing; formulating and checking mathematical claims and proof steps; proposing and refining hypotheses; designing and reviewing experimental methodology; implementing and reviewing analysis and training code; translating between Korean and English; interpreting experimental results; and creating small synthetic diagnostic examples used for mechanism checks. We also used generative AI for literature search and summarization, identifying related work and research gaps, suggesting experimental parameters, drafting and editing manuscript text, organizing the paper, formatting references and tables, and creating or modifying scientific figures. We did not use generative AI to generate or alter the benchmark datasets or their labels, nor as a substitute for the reported empirical measurements. AI-assisted code, mathematical arguments, citations, tables, figures, and numerical claims were checked against executable code, experiment logs, retained source data, and primary literature before inclusion. The authors reviewed all AI-assisted work and take responsibility for the final content of this paper.
\section*{Reproducibility Statement}
The appendix specifies selection contracts, parameter ranges, evaluation protocols, uncertainty calculations, and cost horizons. The supplemental archive includes a reference implementation of the normalized selector, analytical and trajectory-figure scripts, and processed trajectory summaries used in the paper. These materials reproduce the reported selector calculations and document figures; they are not a complete release of every training harness or raw benchmark artifact. Quantities absent from retained records are identified explicitly.
\clearpage
\bibliography{references}
\bibliographystyle{iclr2027_conference}
\clearpage
\appendix
\section{Mathematical properties and implementation details}
\label{app:proofs}
\subsection{Threshold, cardinality, and no-pruning region}
For $q_i=e^{c_i/\tau}/\overline{e^{c/\tau}}$ and $\kappa=1/\rho$, $u_i=0$ is equivalent to $q_i\ge\kappa$. Taking logarithms gives Eq.~\eqref{eq:cutoff}. If $D$ is the removed set, $n=\sum_iq_i\ge\sum_{i\in D}q_i\ge\kappa|D|$, hence $|D|/n\le\rho$. For $0<\rho<1$, at least one example remains; moreover,
\[
\bar u=\frac1n\sum_i[1-\rho q_i]_+\ge1-\rho>0,
\]
so the continuous-weight denominator is nonzero. All retained coefficients are nonnegative. By Jensen,
$\tau\log\overline{e^{c/\tau}}\ge\bar c$; therefore $\max c-\bar c<\tau\log\kappa$ implies $\max c<T(c)$ and no pruning. The stronger-to-check range condition $\max c-\min c<\tau\log\kappa$ is also sufficient. Equal scores give $q_i=1$ and $u_i=1-\rho$, hence $w_i=1$.

These are conditional statements about the score operator. For $\rho\ge1$, a flat population has an empty support. Nonflat populations can still have nonempty support; a failed run cannot be inferred from its parameter alone. Some historical boosting searches deliberately included this larger domain and stopped on degenerate support rather than replacing it with uniform weights. Their performance does not validate the conservative-domain guarantee.

\subsection{Exact mass criterion: what makes pruning stop?}
Let $\pi_i=e^{c_i/\tau}/\sum_j e^{c_j/\tau}$ denote the normalized exponential mass \emph{across examples}, not a predicted class probability. Since $q_i=n\pi_i$ and $\max_i q_i=n/M_\tau$, the following conditions are equivalent:
\begin{equation}
 D=\varnothing
 \iff \max_i q_i<\kappa
 \iff \max_i\pi_i<\frac{\kappa}{n}
 \iff M_\tau>\frac n\kappa=\rho n.
 \label{eq:no_prune_equivalences}
\end{equation}
The inequality is strict because equality produces a zero coefficient and removes at least one maximum. The total $\sum_i\pi_i=1$ (or $\sum_iq_i=n$) cannot itself diagnose concentration. An unnormalized total is not a meaningful absolute trigger either: replacing $c_i$ by $c_i+a$ changes it while leaving the mask unchanged.

For a particular example, the same rule can be written as
\[
 i\in D \iff \sum_j e^{(c_j-c_i)/\tau}\le\rho n.
\]
Thus removal depends on its exponential gaps to \emph{all} candidates. The criterion is closest to multiplicative exceedance of an empirical exponential mean, or equivalently a softmax-concentration threshold. It is not a density-based outlier test, a calibrated tail probability, or a false-discovery-rate guarantee. No inference about label correctness follows from crossing it.

\begin{table}[h]
\caption{Exact illustrative decisions for $n=100$, $\rho=.7$, $\tau=.05$. No-pruning requires $M_\tau>70$. These constructed distributions are not additional training evidence. The fixed-quantile comparator always removes exactly ten rows.}
\centering\small
\begin{tabular}{@{}lrrrr@{}}\toprule
Confidence population & $M_\tau$ & LACS removed & Top-10\% & $c>.74$ \\\midrule
100 at .95 & 100.00 & 0 & 10 & 100 \\
90 at .70; 10 at .71 & 83.69 & 0 & 10 & 0 \\
90 at .70; 10 at .75 & 43.11 & 10 & 10 & 10 \\
40 at .70; 60 at .75 & 74.72 & 0 & 10 & 60 \\
40 at .70; 60 at .80 & 65.41 & 60 & 10 & 60 \\
20 at .70; 80 at .90 & 80.37 & 0 & 10 & 80 \\\bottomrule
\end{tabular}
\label{tab:mass_examples}
\end{table}
The second and third rows preserve confidence ordering but change the gaps. The third and fourth use the same two values but change the upper-cluster mass. The fourth and fifth show why a common high-confidence group is not categorically protected: sufficient separation can still trigger removal below the cardinality ceiling. A mass such as $M_\tau$ is an algebraic control variable, not evidence that the resulting subset is optimal.

\paragraph{Same mean, different shape.}
Let $\beta=1/\tau$.  The empirical log-mean-exp is the scaled cumulant-generating function
\[
\operatorname{LME}_{\beta}(c)=\frac1\beta\log\left(\frac1n\sum_i e^{\beta c_i}\right).
\]
Because log-sum-exp is symmetric and convex, it is Schur-convex: if two finite score vectors have the same arithmetic mean and $c$ majorizes $c'$, then $\operatorname{LME}_{\beta}(c)\ge\operatorname{LME}_{\beta}(c')$ for every $\beta>0$.  Thus a pure mean boundary is identical for same-mean populations, while the LACS boundary can rise when the population becomes more dispersed.  Expanding the empirical cumulant-generating function around $\beta=0$ gives
\[
\operatorname{LME}_{\beta}(c)=\mu+\frac{\beta}{2}\sigma^2+\frac{\beta^2}{6}\kappa_3+\frac{\beta^3}{24}\kappa_4+O(\beta^4),
\]
where $\kappa_m$ are empirical cumulants.  The variance term is always nonnegative; positive third cumulant gives an additional local upper-tail correction.  This is an interpretation of the boundary, not the rule used numerically.

For example, at $\tau=.2$, the populations $(.5,.5,.5,.5)$ and $(.2,.4,.6,.8)$ both have mean $.5$, but their LME values are $.500$ and $.611$.  A deterministic cutoff based only on the mean places the boundary at the same location for both, while the LACS population reference changes.  This should not be paraphrased as ``pointiness'': concentration at the mean lowers LME toward the mean, whereas dispersion and upper-tail shape can raise it.  Nor does a higher LME alone imply less pruning, because the upper scores themselves can move relative to the adaptive boundary; Eq.~\eqref{eq:twolevel} gives the exact two-cluster decision.

\subsection{Invariances, sensitivity, and quantile equivalence}
A common additive score shift multiplies numerator and denominator by the same constant, leaving $q$ unchanged. Replicating every score the same number of times also preserves $q$ and the selected fraction. Neither property means that scores may be arbitrarily recalibrated: a nonlinear monotone map changes gaps and generally changes the cutoff. A confidence quantile, in contrast, depends only on ordering (and a specified tie rule).

For $n>1$, $q_i$ is strictly increasing in $c_i$ for fixed other scores, so $S$ is a confidence lower-level set. If the realized count is $K$ and no score is tied across the boundary, $S$ is precisely the set of the $K$ lowest scores. Thus a matched-$K$ confidence-ranker experiment cannot distinguish LACS's hard support from quantile selection. It is a valid identity check, not an independent accuracy baseline. A fixed quantile over time differs because it sets $K$ in advance.

The threshold is 1-Lipschitz in the infinity norm: if $\|c-c'\|_\infty\le\epsilon$, bounding the exponentials by $e^{\pm\epsilon/\tau}$ yields $|T(c)-T(c')|\le\epsilon$. This bounds threshold motion, not the number of membership flips: many scores can lie close to the boundary. Also,
\[
\frac{\partial T}{\partial c_j}
=\frac{e^{c_j/\tau}}{\sum_i e^{c_i/\tau}}>0.
\]
Raising one extreme score raises the cutoff for the other examples and cannot newly prune an unchanged survivor. It can instead prevent other high-score examples from being pruned. This masking effect is a reason not to claim robust outlier detection from self-normalization.

\subsection{Separation, mass, and parameter roles}
For scores $a$ and $a+\Delta$, with the latter occurring at fraction $r$, factoring out $e^{(a+\Delta)/\tau}$ gives
\begin{equation}
q_H=\frac{1}{r+(1-r)e^{-\Delta/\tau}},\qquad
q_H\ge\kappa\iff r<\rho\ \text{and}\ \frac{\Delta}{\tau}\ge\log\frac{1-r}{\rho-r}.
\label{eq:twolevel}
\end{equation} The low group has $q_L=[(1-r)+re^{\Delta/\tau}]^{-1}\le1$, so it is retained when $\kappa>1$. The high group is removed if and only if $r+(1-r)e^{-\Delta/\tau}\le\rho$. This is possible for finite positive gaps exactly when $r<\rho$, and rearrangement yields the stated boundary.

\begin{table}[h]
\caption{Two-level examples at $\rho=.7$. These are exact operator calculations, not training measurements. A quantile is assumed to remove exactly 10\% with random tie-breaking.}
\centering\small
\begin{tabular}{@{}rrrr@{}}\toprule
\shortstack{High-group\\fraction $r$} & \shortstack{Minimum\\$\Delta/\tau$} & \shortstack{LACS removal\\above boundary} & \shortstack{Quantile recall\\within high group}\\\midrule
.01 & .361 & 1\% & 100\%\\
.10 & .405 & 10\% & 100\%\\
.40 & .693 & 40\% & 25\%\\
.60 & 1.386 & 60\% & 16.7\%\\
.70 or more & no finite boundary & 0\% (two-level case) & $.1/r$\\\bottomrule
\end{tabular}
\end{table}
At $r<.1$, a fixed 10\% quota also removes some bulk; its bulk removal rate is $(.1-r)/(1-r)$. At $r>.1$, it removes only a fraction $.1/r$ of the high group. These formulas avoid treating a tied group's presence in the selected set as 100\% recall.

Increasing $\kappa$ at fixed scores makes pruning more conservative and lowers the ceiling $1/\kappa$. Temperature changes the concentration of normalized exponentials. As $\tau\to\infty$, $q_i\to1$ and the conservative-domain selector retains all examples. As $\tau\to0$, a unique maximum absorbs the normalized mass; with $n>\kappa$, eventually that maximum alone is removed. Intermediate temperatures can therefore remove more examples than either limiting case. A Gaussian-score calculation illustrates, rather than proves universality of, this nonmonotonicity: for an idealized unbounded score population with $a=\sigma/\tau$,
\[
\Pr\{c\ge T\}=1-\Phi\!\left(\frac a2+\frac{\log\kappa}{a}\right),
\]
which is maximized at $a=\sqrt{2\log\kappa}$. At $\kappa=1/.7$ the maximum is approximately .199. Predictive probabilities are bounded, so this Gaussian model is only a shape calculation, not a calibrated probability model or a universal pruning-rate limit.


\begin{figure}[h!]
\centering
\includegraphics[width=.72\linewidth]{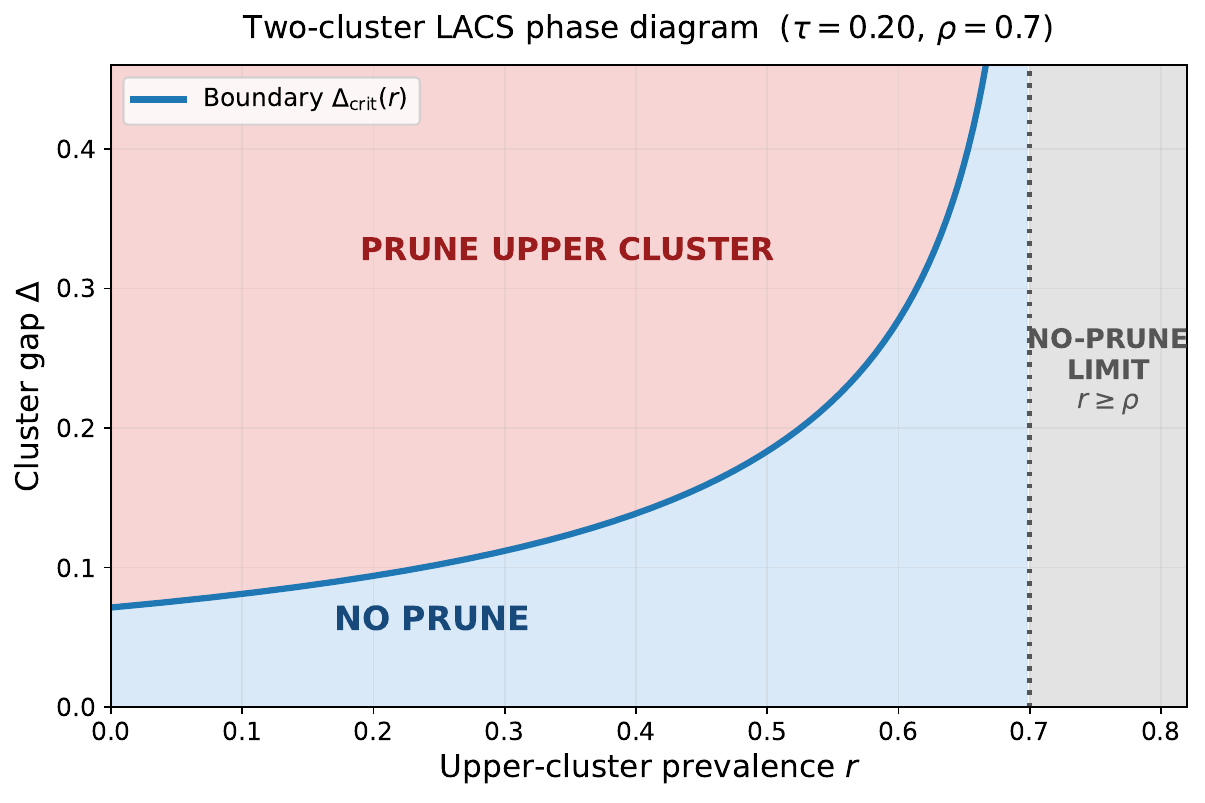}
\caption{\textbf{Illustrative two-cluster phase diagram.}  Above the boundary, LACS prunes the upper cluster; below it, LACS does not prune.  For $r\ge\rho$, the two-cluster model lies in a no-prune limit because a tied upper group of that mass cannot cross the finite-gap cutoff.  The region labels use the same parameters as Figure~\ref{fig:geometry-main}.}
\label{fig:phase-refined}
\end{figure}

\subsection{Exact selective execution and its limits}
At a fixed state, detaching $w$ gives Eq.~\eqref{eq:exact} by dropping zero terms from a finite sum. Its implementation requires unchanged retained activations and Jacobians, original candidate normalization, and matched stochastic realizations. Batch-independent normalization such as LayerNorm or GroupNorm permits this decomposition; BatchNorm changes a retained example's computation when other examples are removed. Augmentation, dropout, contrastive coupling, and batch-level losses require separate care. A fresh scoring forward can also change the state if running statistics or other mutable buffers are updated.

This is an exact identity for the frozen weighted surrogate
$\widetilde L_t(\theta)=n^{-1}\sum_i w_i(\theta_t)\ell_i(\theta)$.
Recomputing and detaching weights every step does not in general equal differentiating $n^{-1}\sum_iw_i(\theta)\ell_i(\theta)$, which includes derivatives through $w$. The neural experiments test the former, not a claimed globally conservative vector field. Different floating-point reduction orders can perturb multi-step trajectories even when same-state gradients agree to numerical precision.

\section{Evaluation protocol and inference}
\label{app:stats}
\subsection{Reporting means and standard deviations}
For a single-dataset experiment with $n_s$ seeds, we report $\bar a\pm s_a$, with the unbiased sample SD $s_a$ (denominator $n_s-1$). Paired inference first forms $d_s=a_s-b_s$ and uses $\bar d\pm t_{.975,n_s-1}s_d/\sqrt{n_s}$. The SD of paired differences is not the difference of the two marginal SDs. Neither $\bar a\pm s_a$ nor its overlap with another method's interval is a significance test.

For multi-dataset studies, let $d_{js}$ be the paired difference in dataset $j$ and split $s$. We first average splits to obtain $\bar d_j$, then compute a macro mean and inference across datasets. In the compact study the independent aggregation unit is 14 dataset means, not 42 resplits; in the widened tree study it is also 14 means, not 140. Raw macro-score SDs describe variation across task difficulty. Appendix tables with one dataset per row instead display within-dataset split SDs. We make no probabilistic claim that this finite collection is a random sample of all tasks. Dataset-level aggregation and multiple-comparison issues have long been emphasized in classifier evaluation \citep{demsar2006comparisons}; our particular $t$ and sign-flip procedures are specified here rather than attributed to that reference.

The exact sign-flip statistic enumerates all $2^{14}$ sign assignments and compares the absolute mean with the observed absolute mean. It requires the corresponding null symmetry/exchangeability assumption. The $t$ procedure instead relies on a reasonable model for the mean paired difference. The two can disagree; both are retained where available rather than selecting the more favorable test. The $n=3$ neural exact sign-flip test has coarse resolution and cannot establish a two-sided $p<.05$ even with all signs favorable.

\subsection{Comparison families and significance markers}
The original compact primary family consists of six LACS--random comparisons. Its Holm-adjusted permutation values are reported with $\ast$. Full-data contrasts were secondary; the added $\dagger$ uses a \emph{post-hoc} Holm correction across the six full-data permutation values. That correction is calculated from the source's four-decimal $p$ values, so its displayed precision is limited. It does not retroactively make the full-data family primary or adjust all historical research choices. Nominal 95\% $t$ intervals remain nominal after a $p$-value correction.

The tree expansion used a pilot-triggered design. Its recorded correction is across three depths on new comparison blocks only. Pooled results that include the initial screen are descriptive; the new-only depth-2 result passes its corrected threshold. Independently tuned GOSS comparisons and expanded-resplit AdaBoost comparisons are secondary. The ten-resplit study expanded from five after intermediate values existed, and prior datasets were reused. Its uncertainty is therefore not described as a fully untouched confirmatory design.

All principal neural comparisons have only three seeds. We do not equate failure to reject a difference with evidence of negligible degradation. A future noninferiority claim would need a prespecified practical margin, appropriate test, and sufficient independent replication. We do not add such a margin after observing these results.

\paragraph{Dataset and architecture sources.}
CIFAR-10/100 originate from \citet{krizhevsky2009tiny}; MNIST, Fashion-MNIST, and SVHN are attributed to \citet{lecun1998document,xiao2017fashion,netzer2011svhn}. The residual, wide-residual, attention, and vision-transformer components follow the architecture families of \citet{he2016resnet,zagoruyko2016wide,vaswani2017attention,dosovitskiy2021vit}. LayerNorm and GroupNorm are attributed to \citet{ba2016layer,wu2018group}. These citations identify data/model origins, not identical preprocessing, pretrained weights, or full original recipes; our variants and splits are specified below.

\subsection{Data, hyperparameters, and reuse}
\label{app:protocols}
\begin{table}[h!]
\caption{\textbf{Which implementation does each result test?} A common confidence-selection principle does not imply identical coefficients or cutoff policies. Guarantees for $0<\rho<1$ apply only in that domain.}
\label{tab:contracts}
\centering\small
\begin{tabularx}{\linewidth}{@{}p{.23\linewidth}p{.26\linewidth}X@{}}\toprule
Setting & Support rule & Retained weights / domain \\midrule
Neural selective training & LACS, minibatch scores & Continuous detached weights; $\rho=.7$ \\
Compact weak learners & LACS, ensemble confidence & Ordinary logistic fitting weights; searched $0<\rho<4$ \\
Tree/GOSS comparisons & LACS-T, fixed $|F|$ cutoff & Unit fitting weights; not the normalized cutoff \\
RBF-SVM case & LACS, ensemble confidence & Native LACS weights; fixed $\rho\simeq.864$ \\
Semi-supervised learning & LACS, cached unlabeled pool & FlexMatch loss; admissible HPO $0<\rho<1$ \\bottomrule
\end{tabularx}
\end{table}
\paragraph{Neural baseline panel.}
CIFAR-10 uses ResNet-32 with GroupNorm(8), 50 epochs, batch 128, SGD with learning rate .1, momentum .9 and weight decay $5\times10^{-4}$, five warmup epochs followed by cosine decay. Validation consists of 2,000 stratified official-test images, and the remaining 8,000 form the reporting test partition. Seed zero is used for validation selection and seeds 1--3 for the frozen panel. Source JSONs also contain intermediate test diagnostics at epochs 10 and 25. The reported endpoint is the fixed final epoch; this is not a claim that the test set was accessed only once. The benchmark is therefore not directly comparable with published 10,000-image CIFAR-10 scores.

LACS fixes $\rho=.7,\tau=.05$. RS2 searches retained fractions $\{.3,.5,.712\}$ and compresses the host recipe's learning-rate schedule over executed steps. RCAP searches the same fractions and $\beta\in\{1,3\}$; the port follows the paper's class allocation and uses loss clipping at the maximum epoch-zero loss. Selective-Backprop searches $\beta\in\{.4,1,2,3\}$, starts with a full epoch, and keys learning rate to cumulative backward examples. Frozen choices are RS2 .712, RCAP .712/$\beta=3$, and SB $\beta=.4$. These adaptations are stated so that method labels do not imply unmodified upstream training pipelines. InfoBatch uses the same base trainer, a seed-zero validation search over retained-easy-pool parameters $\{.3,.5,.7\}$, and a frozen choice of .7 with a final full-data phase ($\delta=.875$). Its realized nominal backward fraction is .8137, not .7. The three InfoBatch seeds were retrieved after the original baseline panel; all full-CE replay values and the available LACS replay values reproduce the earlier frozen accuracy records. The same-host three-seed LACS and CE replays now reproduce the original prediction values. Their completed timing series and separate work-calibrated InfoBatch setting are reported in Appendix~\ref{app:neural_timing}. RePB comparisons use the same data/model/optimizer recipe but a different execution host, as detailed there.

\paragraph{Tree panels.}
The tree cache contains numeric binary derivatives of 14 tabular tasks associated with the benchmark studied by \citet{grinsztajn2022tabular}. The producer reads cached arrays and uses a binary target conversion. Training size is $\lfloor\min(.7n,10000)\rfloor$; 30\% of the remaining rows are validation and the rest test. These are cached derivatives, not a claim to reproduce every original task's published preprocessing or split protocol. Cached transformations and examples may have been accessed before the present split. In the resplit study the same underlying examples are repartitioned, so frozen hyperparameters do not imply an untouched holdout population.

The original tree/GOSS campaign uses 200 shared configuration draws and a 500-round cap. The separate widened depth-one campaign uses 200 draws and a 2,000-round cap. Its recorded shared search is learning rate log-uniform $[10^{-3},3]$, row and feature subsampling uniform $[.5,1]$, minimum child weight rounded log-uniform $[1,100]$, and leaf regularization log-uniform $[1,4]$. LACS-T searches $\vartheta$ log-uniform $[.05,20]$; tuned random searches its retained fraction uniformly in $[.1,1]$. The standard full-data arm may itself use XGBoost's built-in subsampling; ``full'' means no additional LACS filter. Configuration and round are validation-selected. The widened resplit configurations and round counts are then frozen. The AdaBoost family \citep{freund1997decision} supplies the external stump baseline: the implemented SAMME variant uses stump learners, learning rates $\{.1,.5,1\}$ and a 2,000-estimator cap, not the same 200-draw search budget.

\paragraph{Compact nonlinear fitting.}
The 14 datasets are credit, california, bank-marketing, heloc, MiniBooNE-10k, HIGGS-20k, Diabetes130US, electricity, MagicTelescope, house-16H, pol, eye-movements, jannis, and credit-card default. This is not identical to the tree panel: the high-dimensional Bioresponse task is omitted for runtime and HIGGS is included. Size duplicates are not counted as different tasks. Stage one uses split zero; the final panel uses resplits 1--3. Six learner configurations are promoted; MLP width eight was screened but not promoted under the recorded rule.

MLP raw output is $a^\top\tanh(Wx+b)+c$, with $dH+2H+1$ parameters. Quadratic raw output is $b+w^\top x+\sum_{r=1}^{R}a_r(u_r^\top x)^2$, with $d+Rd+R+1$ parameters. Both are followed by $\tanh$. Full-batch Adam fits weighted logistic loss with gradient clipping at 10. The post-smoke grid is learning rate $\{.003,.01,.03\}$, weight decay $\{0,10^{-4},10^{-3}\}$, and 100 or 300 optimizer steps. The initial 10/30-step smoke undertrained MLPs; this grid was amended before screening. Each cell evaluates 32 paired candidates and at most 100 boosting rounds, with shared initialization streams.

In historical experiment logs, the sign convention is $\lambda:=-\rho$; the current manuscript uses $\rho>0$ for the normalized rule.  The inherited selector search draws $\lambda$ uniformly in $(-4,0)$, $\tau$ log-uniform in $[10^{-3},200]$, shrinkage log-uniform in $[10^{-3},1]$, and warmup from $\{0,2,5,10,20\}$. Thus some candidate selectors are outside $0<\rho<1$. Empty supports and degenerate fits are stopped and recorded, not silently changed to all-row fitting. Ordinary logistic fitting weights are normalized over retained rows. The stage coefficient uses the retained-set weighted sign error; it is not asserted to be an exact line search for a real-valued $\tanh$ weak learner. LACS and full training choose their own configuration and round on validation. The random control uses LACS's configuration and per-round counts, but chooses its own validation round. This changes both the fitted weak learner and its coefficient and may yield different selected-prefix budgets.

\section{Neural results and baseline coverage}
\label{app:infobatch}
\subsection{Raw confirmation and work accounting}
\begin{table}[h!]
\caption{\textbf{CIFAR-10/GN direct pruning baselines, three seeds.} Accuracy and available wall times are mean $\pm\sdS$. V: validation-selected; B: backward-matched; N: near-backward budget. Wall times are from the four-arm InfoBatch host only; a dash means no comparable three-seed wall series.}
\label{tab:neural}
\centering\small\begingroup\setlength{\tabcolsep}{3.5pt}
\begin{tabular}{@{}lrrrrr@{}}\toprule
Method & Test acc. $\uparrow$ & Bwd. (M) & Extra fwd. (M) & Steps & Wall (s) $\downarrow$ \\\midrule
Full CE & $89.82\pm0.20$ & 2.496 & 0.000 & 19,500 & $312.9\pm0.5$ \\
LACS & $89.77\pm0.51$ & 1.776 & 2.496 & 19,500 & $396.8\pm3.4$ \\
InfoBatch (V) & $89.75\pm0.21$ & 2.034 & 0.000 & 15,893 & $271.6\pm1.8$ \\
InfoBatch (B) & $89.06\pm0.33$ & 1.776 & 0.000 & 13,876 & $244.5\pm0.8$ \\
RePB (V) & $89.33\pm0.19$ & 2.041 & 0.000 & 15,948 & --- \\
RePB (N) & $89.46\pm0.05$ & 1.827 & 0.000 & 14,274 & --- \\
RCAP & $89.11\pm0.08$ & 1.779 & 0.050 & 13,900 & --- \\
RS2 & $88.90\pm0.12$ & 1.779 & 0.000 & 13,900 & --- \\
Selective-Backprop & $89.55\pm0.14$ & 1.795 & 2.446 & 14,047 & --- \\
\bottomrule
\end{tabular}\endgroup

\end{table}
Table~\ref{tab:seed_raw} preserves all three-seed prediction series. Positive signs across three seeds do not by themselves establish superiority. The new comparisons extend the earlier frozen neural panel; they are not additional independent LACS seeds.
\begin{table}[h]
\caption{CIFAR-10 final test accuracy by seed, fixed validation/test partition and network recipe. V/B/N are defined in Table~\ref{tab:neural}. RePB reuses the InfoBatch-host LACS/CE series; exact cross-host prediction reproduction is checked only at seed 1.}
\label{tab:seed_raw}\centering\small\begingroup\setlength{\tabcolsep}{4pt}
\begin{tabular}{@{}lrrr@{}}\toprule
Method & Seed 1 & Seed 2 & Seed 3 \\\midrule
Full CE & 89.7750 & 89.6500 & 90.0375 \\
LACS & 89.9000 & 90.2000 & 89.2125 \\
InfoBatch (V) & 89.9875 & 89.5875 & 89.6750 \\
InfoBatch (B) & 88.6875 & 89.3250 & 89.1750 \\
RePB (V) & 89.5375 & 89.1750 & 89.2875 \\
RePB (N) & 89.4000 & 89.5000 & 89.4750 \\
RCAP & 89.0625 & 89.2000 & 89.0625 \\
RS2 & 88.8000 & 89.0250 & 88.8625 \\
Selective-Backprop & 89.5125 & 89.4375 & 89.7000 \\
\bottomrule
\end{tabular}\endgroup

\end{table}

\begin{table}[h]
\caption{Paired accuracy uncertainty for the main CIFAR-10 baseline table. Every row is LACS minus the comparator; positive favors LACS. These nominal intervals are kept in the supplement because all include zero.}
\label{tab:neural_ci}\centering\small\begingroup\setlength{\tabcolsep}{4pt}
\begin{tabular}{@{}lrrrr@{}}\toprule
Comparator & Mean $\pm$ SD & 95\% $t$ CI & $p_t$ & $p_{\rm sf}$ \\\midrule
Full CE & $-0.050\pm0.704$ & $[-1.799,+1.699]$ & $0.913$ & $1.000$ \\
InfoBatch (V) & $+0.021\pm0.546$ & $[-1.335,+1.376]$ & $0.953$ & $1.000$ \\
InfoBatch (B) & $+0.708\pm0.605$ & $[-0.794,+2.211]$ & $0.180$ & $0.250$ \\
RePB (V) & $+0.438\pm0.554$ & $[-0.938,+1.813]$ & $0.305$ & $0.500$ \\
RePB (N) & $+0.313\pm0.508$ & $[-0.949,+1.574]$ & $0.398$ & $0.500$ \\
RCAP & $+0.663\pm0.451$ & $[-0.458,+1.783]$ & $0.126$ & $0.250$ \\
RS2 & $+0.875\pm0.456$ & $[-0.258,+2.008]$ & $0.080$ & $0.250$ \\
Selective-Backprop & $+0.221\pm0.641$ & $[-1.373,+1.814]$ & $0.611$ & $0.750$ \\
\bottomrule
\end{tabular}\endgroup

\end{table}

The nominal stream has 50,000 images over 50 epochs (2.5 M rows), but dropping incomplete \emph{training} batches yields 2.496 M full-training rows. LACS's nominal backward fraction .7102 is .7114 relative to actual full training. Absolute counters avoid this ambiguity. LACS has 2.496 M extra scoring-forward rows; RCAP has .050 M, RS2 zero, and SB 2.446 M. Training-forward rows equal backward rows in the four-arm InfoBatch panel. Matching backward counts does not match these forwards or optimizer steps.

\paragraph{Operating-point selection.}
The InfoBatch port uses the official class at commit \texttt{8bbb012}. Seed-0 validation selects easy-pool keep .7 from $\{.3,.5,.7\}$, with final full-data fraction governed by $\delta=.875$. A separate keep value .537 is calibrated using backward counts only, not accuracy, to target LACS's nominal .7102 fraction. At reporting seeds, it executes 1,776,085 backward rows on average versus LACS's 1,775,604, a difference of approximately .027\% of LACS's count. This is approximate count matching, not exact per-step support replay.

RePB uses the upstream \texttt{repb.py} at commit \texttt{80e217d}, with $\alpha=B=128$. Validation selects keep .7 from $\{.25,.4,.55,.7\}$; interpolation of pilot work selects .509 for the near-budget run. Its observed .7308 nominal fraction is above LACS's .7102; this is explicitly \emph{not} exact matching. The on-host freeze file was not recovered; the choices were reconstructed from the seed-0 pilot JSONs and match those executed. Our port preserves the upstream code's first-epoch prefix, $(E+1)/N$ weights, fixed-count resampling, dropped sub-window tail, and per-epoch pruning reseeding. It uses ResNet-32/GN, 50 epochs, the common SGD recipe, and epoch-progress LR rather than the original full model/optimizer/schedule. This is a code-faithful common-recipe comparison, not a complete reproduction of the published recipe.

\subsection{Completed same-host runtime comparisons}
\label{app:neural_timing}
All four InfoBatch-panel methods run on one host with two RTX 3090 GPUs, one job per GPU at a time. The GPUs run concurrently and share host resources; run order and initial warm-up can affect elapsed time. We retain the complete measured series (Table~\ref{tab:neural_wall_raw}) without converting backward savings into total-FLOP savings. Ratios of mean wall time are 1.268 for LACS/full CE, 1.461 for LACS/InfoBatch-V, and 1.623 for LACS/InfoBatch-B. These results do not support a LACS speedup in this setting.
\begin{table}[h]
\caption{InfoBatch-host wall times (seconds). Means and sample SD use the three measured runs; this series replaces the incomplete timing snapshot in the earlier manuscript.}
\label{tab:neural_wall_raw}\centering\small\begingroup\setlength{\tabcolsep}{4pt}
\begin{tabular}{@{}lrrrr@{}}\toprule
Method & Seed 1 & Seed 2 & Seed 3 & Mean $\pm$ SD \\\midrule
Full CE & 312.7 & 313.5 & 312.5 & $312.9\pm0.5$ \\
LACS & 397.8 & 393.0 & 399.5 & $396.8\pm3.4$ \\
InfoBatch (V) & 271.4 & 269.9 & 273.5 & $271.6\pm1.8$ \\
InfoBatch (B) & 244.2 & 245.4 & 244.0 & $244.5\pm0.8$ \\
\bottomrule
\end{tabular}\endgroup

\end{table}

RePB runs on a separate two-RTX-3090 host. Its three-seed accuracy comparisons reuse LACS and CE from the InfoBatch host. Seed-1 accuracy is reproduced exactly for both reference methods, but no analogous seed-2/3 check is recorded. Only seed 1 has the complete same-host reference timing panel below. It is not pooled with Table~\ref{tab:neural_wall_raw}, and no single-run SD is fabricated.
\begin{table}[h]
\caption{Separate RePB-host timing panel: seed 1 only, same CIFAR-10 recipe. This is a measured case, not a three-seed timing comparison.}
\label{tab:repb_seed1_wall}\centering\small\begingroup\setlength{\tabcolsep}{4pt}
\begin{tabular}{@{}lr@{}}\toprule
Method & Seed-1 wall time (s) \\\midrule
Full CE & 343.1 \\
LACS & 397.1 \\
RePB (V) & 306.7 \\
RePB (N) & 274.5 \\
\bottomrule
\end{tabular}\endgroup

\end{table}

The eight neural accuracy contrasts in Table~\ref{tab:neural_ci} are nominal paired intervals with the dataset fixed. The original five-contrast post-hoc Holm values remain in the accompanying statistical export. The three newly added contrasts (InfoBatch-B, RePB-V, RePB-N) are exploratory and carry no retrospective preregistration. All eight nominal intervals include zero; neither extra comparisons nor multiple-testing corrections create a supported superiority claim.

\subsection{InfoBatch comparisons already performed}
A historical balanced CIFAR-100/BatchNorm panel gives the single-run values in Table~\ref{tab:ib_balanced}. It is retained as evidence rather than replaced by a positive long-tail comparison. Its older LACS implementation differs from the newer CIFAR-10/GN panel.
\begin{table}[h]
\caption{Historical balanced CIFAR-100 comparison, one seed. Different network/training protocol from Table~\ref{tab:neural}. No SD or significance is available. Runtimes are the recorded panel values, not an optimized universal method comparison.}
\label{tab:ib_balanced}\centering\small
\begin{tabular}{@{}lrr@{}}\toprule
Method & Test accuracy (\%) & Recorded wall (s)\\\midrule
Full CE & 78.77 & 2608.30\\
Native InfoBatch-CE & 78.30 & 2121.12\\
Older LACS subset-BN & 78.06 & 3280.53\\
Selective-Backprop & 78.22 & 3358.99\\\bottomrule
\end{tabular}
\end{table}
A separate CIFAR-100-LT experiment uses Focal loss and a final 25-epoch full-data phase. The six paired seed values produce Table~\ref{tab:ib_focal}; the source reports a LACS--InfoBatch difference of $+1.237$ pp with 95\% CI $[-.037,2.510]$, and LACS--dense Focal $-1.085$ pp $[-1.909,-.261]$. LACS uses more rows, updates, and time than InfoBatch. It cannot substantiate simultaneous accuracy and efficiency superiority. Small differences between a mean recomputed from two-decimal seed values and the higher-precision reported mean are rounding, not extra runs.
\begin{table}[h]
\caption{Six-seed Focal comparison: mean $\pm\sdS$ from the recorded per-seed values. The distinct Focal objective prevents merging this table with CE results.}
\label{tab:ib_focal}\centering\small\begin{tabular}{@{}lrrrr@{}}\toprule
Method & Test acc. & Bwd. rows (M) & Updates & Wall (s) \\\midrule
Cached LACS + full tail & $39.50\pm0.79$ & 1.562 & 17,000 & 756 \\
InfoBatch-Focal & $38.26\pm0.66$ & 1.515 & 11,934 & 636 \\
Random + full tail & $39.06\pm0.44$ & 1.562 & 17,000 & 767 \\
Dense Focal & $40.59\pm0.31$ & 2.169 & 17,000 & 736 \\
\bottomrule
\end{tabular}

\end{table}
The existing implementation fidelity checks compare InfoBatch sampling indices, retained-example compensation weights, losses, gradients, and the transition to final full-data training against the reference package. Passing those gates verifies the tested implementation, not every later port or an accuracy advantage. The new CIFAR-10 comparison in Table~\ref{tab:neural} is an additional same-harness test, not a replacement for the historical evidence above. It now includes both InfoBatch operating points, the same-host rerun times, and completed RePB common-recipe comparisons. Their protocol limitations remain attached; completion does not establish a statistical advantage. The historical panels above are retained under their original losses, model choices, and cost definitions.

\subsection{Update-matched, CIFAR-100, and label-aware controls}
\label{app:controls}
All runs use ResNet-32 with GroupNorm, 50 epochs, an 8,000-image test partition and a 2,000-image validation partition. Wall times are same-host and are not comparable across panels.

\begin{table}[h]
\caption{Update-matched control on CIFAR-10, five seeds, one host per seed. Small-batch CE uses batch 91 for 19,764 steps; random drop keeps a uniformly random subset of the LACS support size. Paired 95\% $t$ intervals.}
\label{tab:smallbatch}\centering\small
\begin{tabular}{@{}lrrrr@{}}\toprule
Arm & Test acc. (\%) & Bwd. frac. & Steps & Wall (s)\\\midrule
Full CE & $89.657\pm0.287$ & .998 & 19,500 & 289\\
LACS & $89.950\pm0.438$ & .709 & 19,500 & 280\\
Small-batch CE (91) & $88.945\pm0.277$ & .719 & 19,764 & 288\\
Random drop & $88.765\pm0.276$ & .693 & 19,500 & 381\\\midrule
LACS $-$ small-batch CE & \multicolumn{4}{l}{$+1.005$ $[+.572,+1.438]$ \quad per seed $+1.48,+.79,+.56,+1.05,+1.15$}\\
LACS $-$ random drop & \multicolumn{4}{l}{$+1.185$ $[+.665,+1.705]$ \quad per seed $+1.16,+.75,+.82,+1.75,+1.44$}\\
LACS $-$ full CE & \multicolumn{4}{l}{$+.292$ $[-.404,+.989]$}\\
Small-batch CE $-$ full CE & \multicolumn{4}{l}{$-.713$ $[-1.002,-.423]$}\\\bottomrule
\end{tabular}
\end{table}

\begin{table}[h]
\caption{CIFAR-100 head-to-head, three seeds, one host per seed; recipes frozen from seed-0 validation and work counters. Paired 95\% $t$ intervals for LACS minus each budget-matched arm.}
\label{tab:c100}\centering\small
\begin{tabular}{@{}lrrrl@{}}\toprule
Arm & Test acc. (\%) & Bwd. frac. & Wall (s) & LACS $-$ arm\\\midrule
Full CE & $62.41\pm0.31$ & .998 & 203 & \\
LACS & $62.24\pm0.46$ & .870 & 264 & \\
InfoBatch, work-matched & $61.77\pm0.25$ & .868 & 190 & $+.47$ $[-1.13,+2.07]$\\
RePB, near budget & -- & -- & -- & $+.99$ $[-.46,+2.44]$\\
Selective-Backprop, work-matched & $61.76\pm0.80$ & .863 & 254 & $+.48$ $[-2.29,+3.24]$\\
RS2 & $61.27\pm0.48$ & .863 & 179 & $+.96$ $[+.46,+1.46]$\\\bottomrule
\end{tabular}
\end{table}

\begin{table}[h]
\caption{Label-aware diagnostic on CIFAR-10, three seeds on one host. InfoBatch keep .556 is calibrated on seed-0 backward counters only. C-LACS is a diagnostic, not the proposed method.}
\label{tab:clacs}\centering\small
\begin{tabular}{@{}lrrr@{}}\toprule
Arm & Test acc. (\%) & Bwd. frac. & Wall (s)\\\midrule
C-LACS & $90.150\pm0.283$ & .721 & 226\\
LACS & $89.742\pm0.460$ & .710 & 221\\
InfoBatch, work-matched & $89.254\pm0.231$ & .722 & 139\\\midrule
C-LACS $-$ InfoBatch & \multicolumn{3}{l}{$+.896$ $[+.119,+1.672]$}\\
LACS $-$ InfoBatch & \multicolumn{3}{l}{$+.487$ $[-1.222,+2.197]$}\\\bottomrule
\end{tabular}
\end{table}

\paragraph{Cached (lagged) confidences: a negative result.}
Replacing the fresh scoring forward with stored confidences does not make LACS faster on this setup and lowers accuracy. In the five-seed panel of Table~\ref{tab:smallbatch}, an EMA-lagged score keeps more rows (backward fraction .839, 352 s) and is $-.312$ pp from LACS ($[-1.107,+.482]$); a purely lagged score keeps fewer rows (.647, 341 s) and is $-.892$ pp ($[-1.376,-.409]$). At ResNet-32/CIFAR scale the scoring forward is not the wall-clock bottleneck.

\paragraph{Default-ratio execution matrix with CE and InfoBatch.}
Six configurations compare LACS (frozen $\lambda=-.7$, $\tau=.05$) with CE and InfoBatch at its default prune ratio, three seeds each. These are native-setting comparisons, not tuned or work-matched results, and InfoBatch does less backward work in five of six configurations. LACS minus CE intervals include zero in every configuration. LACS minus InfoBatch is $+.68$ pp $[+.32,+1.04]$ on CIFAR-10/ViT-Tiny, $+2.21$ $[+.73,+3.70]$ on CIFAR-100/ResNet-32 and $+3.48$ $[+.74,+6.22]$ on CIFAR-100-LT/ResNet-32, and is unresolved on FashionMNIST/MLP, SVHN/ResNet-32 and CIFAR-100/ViT-Tiny. Work-matched comparisons are Tables~\ref{tab:c100} and~\ref{tab:clacs} and are not pooled with this matrix.

\subsection{RBF-SVM replication}
\label{app:svmrep}
Three datasets and up to three splits (eight cells) repeat the credit replay protocol with a 300-round cap on one 32-core CPU host; absolute times are inflated by contention, so only within-cell ratios are compared. Early stopping (weighted error $\ge.5$) ends several arms before 300 rounds, so errors are at the last executed round. Three cells (california s2, credit s0, credit s1) have all arms running 300 rounds with real pruning; there LACS reduces wall time by 30.2, 37.6 and 45.3\% relative to full fitting and changes it by $+26.5$, $+6.5$ and $-.2\%$ relative to a same-size random support. In bank-marketing s1 the LACS and random arms stop at 267 rounds versus 300 for full fitting; LACS reduces wall time by 71.1\% (67.5\% per executed round) and is 15.3\% faster than random. In bank-marketing s0, bank-marketing s2 and credit s2 the pruned arms stop after 23, 37 and 76 rounds versus 258--300, so their 89--94\% savings mainly reflect earlier stopping; per-round comparisons over such different round ranges are not reported as speed evidence. One cell (california s0) prunes nothing. Final errors relative to full fitting range from $-1.96$ to $+.57$ pp, and credit s0 does not reproduce its stored validation prefix on this host and is flagged.

\section{Compact weak learners: full results and cost horizons}
\label{app:compact}
\begin{table}[h]
\caption{All six promoted compact learners against random supports at the LACS configuration. Error is mean $\pm\sdD$; LACS errors are in Table~\ref{tab:compact}. $\ast$ denotes the recorded primary-family Holm correction $p<.05$. Random replays per-round counts but chooses its own validation round.}
\centering\small\begin{tabular}{@{}lrrrrrr@{}}\toprule
Learner & Random error & $\Delta_{\rm err}$ & 95\% $t$ CI & $p_{\rm sf}$ & $p_{\rm H}$ & Wins \\\midrule
MLP H1 & $24.15\pm11.30$ & $-0.169$ & $[-0.357,+0.020]$ & $0.052$ & $0.104$ & 10/14 \\
MLP H2 & $22.95\pm11.40$ & $-0.223^{\ast}$ & $[-0.418,-0.028]$ & $0.005$ & $0.025$ & 11/14 \\
MLP H4 & $22.24\pm11.42$ & $-0.119$ & $[-0.271,+0.032]$ & $0.112$ & $0.112$ & 8/14 \\
Quad R1 & $23.94\pm11.31$ & $-0.601^{\ast}$ & $[-1.097,-0.104]$ & $0.011$ & $0.037$ & 10/14 \\
Quad R2 & $23.58\pm11.27$ & $-0.621^{\ast}$ & $[-1.065,-0.176]$ & $<.001$ & $0.001$ & 13/14 \\
Quad R4 & $23.40\pm11.12$ & $-0.584^{\ast}$ & $[-1.135,-0.033]$ & $0.009$ & $0.037$ & 9/14 \\
\bottomrule
\end{tabular}

\end{table}
\begin{figure}[h]
\centering\includegraphics[width=.84\linewidth]{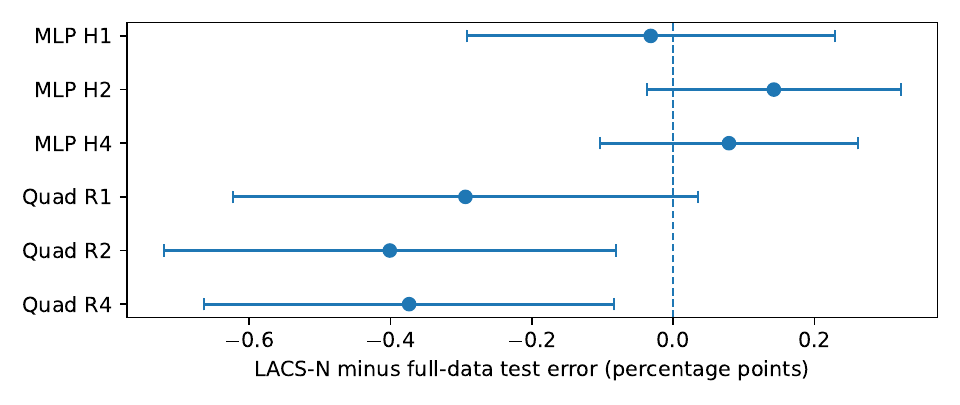}
\caption{Paired macro error differences against full-data fitting. Bars are nominal 95\% $t$ confidence intervals across 14 dataset means, \emph{not SDs}. Both positive and unresolved learner configurations are retained.}
\end{figure}
\begin{table}[h]
\caption{Quadratic R2, per-dataset test error mean $\pm$ SD across three resplits. The macro SD in the main table measures different variation. Neither the large random spread on california nor its favorable LACS difference establishes a universal mechanism.}
\label{tab:quad_ds}\centering\small\begin{tabular}{@{}lrrrr@{}}\toprule
Dataset & Full & LACS & Random & LACS $-$ Full \\\midrule
Diabetes130US & $39.02\pm0.32$ & $39.05\pm0.26$ & $39.39\pm0.44$ & $+0.029$ \\
MagicTelescope & $16.52\pm0.89$ & $16.49\pm1.07$ & $16.69\pm1.08$ & $-0.036$ \\
MiniBooNE\_10k & $7.90\pm0.88$ & $7.92\pm0.64$ & $7.79\pm0.71$ & $+0.016$ \\
bank-marketing & $23.22\pm0.97$ & $22.02\pm0.52$ & $22.71\pm0.79$ & $-1.200$ \\
california & $15.26\pm0.20$ & $15.13\pm0.36$ & $18.03\pm4.59$ & $-0.130$ \\
credit & $25.49\pm0.06$ & $25.07\pm0.42$ & $25.70\pm0.19$ & $-0.418$ \\
credit-card-default & $29.51\pm0.20$ & $29.39\pm0.23$ & $29.73\pm0.09$ & $-0.120$ \\
electricity & $24.70\pm0.28$ & $24.49\pm0.44$ & $24.74\pm0.20$ & $-0.205$ \\
eye\_movements & $43.15\pm2.27$ & $42.30\pm1.38$ & $42.67\pm1.73$ & $-0.855$ \\
heloc & $28.19\pm0.85$ & $28.21\pm1.18$ & $28.35\pm1.52$ & $+0.016$ \\
higgs\_20k & $34.03\pm0.42$ & $32.19\pm0.74$ & $33.82\pm0.99$ & $-1.848$ \\
house\_16H & $14.23\pm0.86$ & $13.63\pm0.53$ & $13.98\pm1.00$ & $-0.600$ \\
jannis & $23.80\pm0.51$ & $23.61\pm0.57$ & $24.06\pm0.52$ & $-0.186$ \\
pol & $2.06\pm0.14$ & $1.98\pm0.12$ & $2.49\pm0.41$ & $-0.079$ \\
\bottomrule
\end{tabular}

\end{table}

\begin{table}[h]
\caption{\textbf{Do not mix cost horizons.} LACS/full ratios; ``mean'' is the mean of run ratios, ``pooled'' the ratio of summed work. Prefix ends at each method's selected predictor. Replay spans all recorded rounds. Selected-prefix time is unavailable for every learner.}
\label{tab:costs}\centering\small
\begin{tabular}{@{}lrrrrrr@{}}\toprule
 & \multicolumn{2}{c}{Selected-prefix rows} & \multicolumn{2}{c}{Replay rows} & \multicolumn{2}{c}{Replay fit time}\\
Learner & Mean & Pooled & Mean & Pooled & Mean & Pooled\\\midrule
MLP H1 & 4.827 & .963 & .887 & .865 & 1.094 & .937\\
MLP H2 & 1.097 & .799 & .821 & .816 & 1.002 & .807\\
MLP H4 & 1.372 & .832 & .788 & .785 & .817 & .723\\
Quad R1 & 4.601 & .776 & .760 & .750 & 1.086 & .892\\
Quad R2 & .936 & .641 & .700 & .695 & .941 & .747\\
Quad R4 & 2.671 & .760 & .663 & .666 & .921 & .678\\\bottomrule
\end{tabular}
\end{table}
Averaging ratios can be dominated by runs where the reference chooses a very early checkpoint. Pooling instead weights expensive or large runs more heavily. Neither is automatically the right single headline; both are reported. For R2, selected rounds average 58.2 for full, 54.2 for LACS, and 53.1 for random. Random and LACS choose the same round in only 19\% of runs. The originally reported 30\% row reduction and .941 fit-time ratio describe full replays, not identical selected endpoints. We do not estimate missing prefix timing by multiplying total runtime by the selected-round fraction.

\section{Tree comparisons and resplit evaluation}
\label{app:trees}
\subsection{GOSS semantics and budget controls}
\begin{table}[h!]
\caption{\textbf{Aggregate tree results across 14 datasets.} Error is mean $\pm\sdD$. LACS-T is the historical fixed unsigned-margin variant, not normalized LACS.}
\label{tab:goss_raw}
\centering\small\begin{tabular}{@{}rrrrr@{}}\toprule
Depth & Full (partner) & GOSS (own HP) & GOSS (matched $K_t$) & LACS-T \\\midrule
1 & $21.74\pm10.47$ & $21.94\pm10.75$ & $23.14\pm10.60$ & $21.50\pm10.74$ \\
2 & $20.45\pm10.99$ & $20.66\pm11.21$ & $20.61\pm11.00$ & $20.27\pm11.01$ \\
3 & $19.98\pm10.79$ & $20.16\pm11.42$ & $20.21\pm10.81$ & $19.99\pm10.95$ \\
\bottomrule
\end{tabular}

\end{table}
\begin{table}[h!]
\caption{\textbf{GOSS paired uncertainty and replication scope.} Negative differences favor LACS-T. New-only Holm values apply to the primary matched-GOSS contrasts across three depths.}
\label{tab:goss_stats}
\centering\small
\begin{tabular}{@{}crrrrr@{}}\toprule
Depth & Matched $\Delta$ & Pooled 95\% CI & $p_{\rm sf}$ & New-only $p_{\rm H}$ & Own-HP $\Delta$ ($p_{\rm sf}$)\\\midrule
1 & $-1.643$ & $[-3.210,-.076]$ & .0049 & .1250 & $-.442$ (.0665)\\
2 & $-.346$ & $[-.604,-.088]$ & .0049 & $.0147^{\ast}$ & $-.386$ (.0314)\\
3 & $-.220$ & $[-.396,-.045]$ & .0068 & .1250 & $-.171$ (.6238)\\\bottomrule
\end{tabular}
\end{table}
Matched GOSS selects approximately two thirds of $K_t$ by the pinned implementation's $|g_ih_i|$ ranking and draws the remainder uniformly, scaling sampled gradients and Hessians by $(n-a_t)/(K_t-a_t)$. Here $a_t$ is the top-group size, not a global pruning fraction. It uses exactly the LACS replay's per-round counts, omits an independent warmup, and executes a global rather than thread-block selection. These are declared adaptations for the matched-support experiment. The all-row identity gate checks that $K_t=n$ reproduces the all-row model; the half-row perturbation must change it. Replays are recomputed together when host-dependent differences affect the LACS count trajectory.

Independently tuned GOSS instead uses top-rate $\{.05,.1,.2,.3\}$ and other-rate $\{.05,.1,.2\}$, 200 shared configuration candidates and its own validation-selected round, with $\lfloor1/\eta\rfloor$ warmup rounds. Official LightGBM GBDT and GOSS references are reported separately because library execution, defaults, and sampling differ. The pinned gradient--Hessian product need not behave like raw-gradient magnitude; in particular, small Hessians can suppress both extreme tails. The experiment is not evidence that every version of GOSS must keep every confident error.

\begin{table}[h]
\caption{Depth-2 raw errors on the 14 datasets. Each row uses one dataset-specific split; randomized controls average their replays. No per-dataset SD is invented for the deterministic single-split arms.}
\centering\small\begin{tabular}{@{}lrrrrr@{}}\toprule
Dataset & LACS-T & Full & GOSS matched & GOSS own & Random \\\midrule
electricity & 15.508 & 16.165 & 15.790 & 16.336 & 15.989 \\
house\_16H & 11.154 & 11.401 & 11.218 & 11.543 & 11.295 \\
pol & 2.408 & 2.502 & 2.446 & 2.172 & 2.351 \\
eye\_movements & 37.523 & 37.523 & 37.523 & 39.337 & 37.523 \\
credit & 22.702 & 22.979 & 22.877 & 22.532 & 22.911 \\
MagicTelescope & 13.843 & 14.235 & 14.285 & 14.093 & 14.292 \\
bank-marketing & 20.117 & 20.072 & 21.440 & 19.802 & 20.765 \\
MiniBooNE & 6.889 & 6.875 & 6.870 & 6.937 & 6.918 \\
Bioresponse & 23.130 & 23.130 & 23.130 & 24.654 & 23.130 \\
credit-card-default & 28.659 & 29.053 & 29.641 & 28.874 & 29.254 \\
jannis & 23.566 & 24.176 & 24.324 & 23.704 & 24.603 \\
Diabetes130US & 39.658 & 39.637 & 39.635 & 39.642 & 39.618 \\
heloc & 28.238 & 28.238 & 28.238 & 28.429 & 28.238 \\
california & 10.371 & 10.357 & 11.188 & 11.123 & 11.093 \\
\bottomrule
\end{tabular}

\end{table}
The primary matched contrast is favorable at every depth in the pooled panel. The initial stage includes six datasets at depths one and three, so those pooled results partly reuse selection-triggering evidence. The new-only comparison has eight additional datasets at those depths and all 14 depth-two blocks. Its primary means are $-2.021$ pp at depth one (CI $[-4.990,.947]$), $-.346$ at depth two ($[-.604,-.087]$), and $-.307$ at depth three ($[-.617,.002]$). Holm-adjusted values are .125, .0147, .125. These are new experimental comparisons on existing data splits, not three independent new-data confirmations.

The independently tuned depth-two GOSS difference is $-.386$ pp, CI $[-.754,-.019]$, nominal sign-flip $p=.0314$. Its counterparts at depths one and three are unresolved. Against full training at the LACS configuration, differences are $-.239$, $-.184$, and $+.008$ pp. At depth two, pooled fitting rows fall to .767 of full, but pooled fitting time rises to 1.529. The tree evidence therefore does not establish that LACS is faster than an optimized XGBoost or LightGBM implementation.

\subsection{Widened depth-one experiment with ten resplits}
\label{app:fresh}
\begin{table}[h]
\caption{Widened depth-one study: test error mean $\pm$ SD across ten resplits of each cached dataset. All hyperparameters and round counts are frozen from prior development. This is distinct from the 500-round GOSS experiment.}
\centering\small\begin{tabular}{@{}lrrr@{}}\toprule
Dataset & LACS-T & Tuned random & Full \\\midrule
Bioresponse & $25.22\pm1.53$ & $28.77\pm1.35$ & $28.03\pm1.61$ \\
Diabetes130US & $39.70\pm0.23$ & $39.56\pm0.29$ & $39.70\pm0.23$ \\
MagicTelescope & $17.62\pm0.48$ & $17.26\pm0.33$ & $17.62\pm0.48$ \\
MiniBooNE & $7.94\pm0.11$ & $7.89\pm0.10$ & $7.93\pm0.10$ \\
bank-marketing & $20.13\pm0.49$ & $21.40\pm0.67$ & $20.36\pm0.78$ \\
california & $11.37\pm0.27$ & $11.75\pm0.35$ & $11.55\pm0.28$ \\
credit & $23.99\pm0.29$ & $23.31\pm0.57$ & $23.51\pm0.53$ \\
credit-card-default & $29.48\pm0.82$ & $29.88\pm0.68$ & $29.45\pm0.88$ \\
electricity & $16.59\pm0.26$ & $17.48\pm0.29$ & $17.00\pm0.30$ \\
eye\_movements & $39.82\pm1.49$ & $40.51\pm1.04$ & $41.46\pm1.86$ \\
heloc & $28.75\pm1.15$ & $28.60\pm1.20$ & $28.75\pm1.15$ \\
house\_16H & $12.71\pm0.27$ & $12.97\pm0.22$ & $12.75\pm0.31$ \\
jannis & $24.75\pm0.17$ & $24.94\pm0.11$ & $24.98\pm0.16$ \\
pol & $4.25\pm0.39$ & $5.58\pm0.42$ & $5.98\pm0.44$ \\
\bottomrule
\end{tabular}

\end{table}
The macro errors are $21.595\pm10.800$ for LACS-T, $22.134\pm10.782$ for tuned random, and $22.078\pm10.872$ for full XGBoost, where these SDs are across dataset means. LACS--random is $-.540$ pp (95\% $t$ CI $[-1.142,.063]$, $p_t=.0750$, $p_{\rm sf}=.0465$). LACS--full is $-.484$ pp ($[-1.011,.044]$, $p_t=.0693$, $p_{\rm sf}=.0645$). A later AdaBoost-stump comparison gives $-.535$ pp ($[-1.028,-.041]$, $p_t=.0358$, $p_{\rm sf}=.0251$), a nominal result rather than a family-adjusted confirmation. The five-to-ten expansion and overlap with development examples are disclosed in Appendix~\ref{app:stats}. The main claim does not depend on choosing the one inferential procedure that excludes zero.

\section{Neural execution and the Transformer case}
\label{app:execution}
\begin{table}[h]
\caption{Dense-weighted versus selective LACS execution. Accuracies are mean $\pm$ seed SD. Standard cells use seeds 1--3; the separate 200-epoch extension uses seeds 0--2. Work reductions are the recorded campaign aggregates; the standard aggregates also include development-seed counters. Time ratios compare selective with dense weighted LACS.}
\centering\small\begin{tabular}{@{}llrrrr@{}}\toprule
Dataset / epochs & Model & Dense LACS acc. & Selective acc. & Bwd. saved & Time ratio \\\midrule
CIFAR-10 / 50 & ResNet-32 GN & $89.64\pm0.14$ & $89.82\pm0.04$ & 28.8\% & 1.192 \\
CIFAR-100 / 50 & ResNet-32 GN & $62.00\pm0.41$ & $62.38\pm0.56$ & 12.9\% & 1.105 \\
CIFAR-100-LT / 50 & ResNet-32 GN & $24.48\pm1.44$ & $24.48\pm0.79$ & 9.9\% & 1.263 \\
SVHN / 50 & ResNet-32 GN & $96.53\pm0.07$ & $96.68\pm0.08$ & 14.1\% & 1.123 \\
CIFAR-10 / 50 & ViT-Tiny LN & $82.40\pm0.10$ & $83.00\pm0.15$ & 20.0\% & 1.149 \\
CIFAR-100 / 50 & ViT-Tiny LN & $56.66\pm0.47$ & $56.48\pm0.36$ & 12.9\% & 1.161 \\
FashionMNIST / 20 & MLP LN & $90.59\pm0.20$ & $90.48\pm0.09$ & 35.1\% & 1.040 \\
CIFAR-100-LT / 200 & ResNet-32 GN & $31.94\pm0.56$ & $32.22\pm0.38$ & 25.5\% & 0.998 \\
\bottomrule
\end{tabular}

\end{table}
The comparison tests execution of one weighting rule rather than superiority over CE; CE accuracies for these configurations were not run in this campaign, so the matrix supports the execution identity and the work counters, not a CE-relative accuracy claim. The matrix and the baseline panel of Table~\ref{tab:neural} are distinct campaigns reporting the same nominal CIFAR-10 architecture and epoch budget. Their selective-LACS means are $89.82$ and $89.77\%$, respectively; aggregate values alone do not establish that the full execution recipes are identical or identify the source of the discrepancy. They are not pooled as paired repetitions. The standard matrix's maximum same-state fp64 relative discrepancy is $3.663\times10^{-15}$. Initial fp32 checks did not all pass the original narrow threshold; the protocol was amended to a fp64 identity gate, while fp32 training results remained separate. No claim of exact trajectory agreement in practical floating point follows. The 200-epoch local extension is not pooled as another independent architecture.

\begin{table}[h]
\caption{Transformer decoder, native LayerNorm and optimized attention, 800 steps, seeds 11--13 on an RTX 4090. All entries are mean $\pm$ sample SD. Quality is validation NLL, not held-out test perplexity or a convergence result.}
\centering\small\begin{tabular}{@{}lrrr@{}}\toprule
Implementation & Wall (s) $\downarrow$ & Val. NLL $\downarrow$ & Bwd. sequences \\\midrule
Optimized CE & $103.91\pm0.56$ & $1.5200\pm0.0093$ & $102400.0\pm0.0$ \\
Optimized dense LACS & $105.56\pm1.20$ & $1.5666\pm0.0092$ & $102400.0\pm0.0$ \\
Custom engine, keep all & $118.67\pm0.96$ & $1.5670\pm0.0073$ & $102400.0\pm0.0$ \\
Custom engine, selective & $110.35\pm0.71$ & $1.5688\pm0.0084$ & $90657.0\pm83.2$ \\
\bottomrule
\end{tabular}

\end{table}
The model has four layers, hidden width 512, eight heads, batch 128 and sequence length 256. The corpus is a byte-level derivative of WikiText-2 \citep{merity2017pointer}; this is not the standard word-level benchmark. Optimized dense LACS and selected custom execution have mean NLL 1.5666 and 1.5688. Selected execution reduces backward sequence count by 11.47\%, but its wall ratio to optimized dense is 1.0454. The ratio to the custom keep-all engine is .9299. The former is the practical same-objective comparison; the latter isolates the gain inside that engine. Earlier measurements using different timing and quality configurations are not substituted for this matched experiment.

\paragraph{SVM details.}
The recorded credit configuration is $C=45.7802$, $\gamma=.106545$, shrinkage $.0084634$, no warmup, $\lambda=-.863959$, and $\tau=.0101386$. The final replay runs 300 rounds. LACS fitting weights differ from the logistic weights in the all-row and random arms; this is a combined-method case. The LACS replay records approximately 1151.5 seconds fitting, 983.3 seconds scoring, and .19 seconds in mask computation. Full pipeline cost is dominated by more than the mask. Summed support-vector incidences cannot be interpreted as the number of unique vectors stored in the final ensemble. Other explored configurations can prune no examples, so the one speed case is not a general SVM acceleration guarantee.

\section{Gradient fidelity, sign, and cutoff controls}
\label{app:gradient}
For binary cross-entropy, let $g_i=\nabla_\theta\ell_i$, $G=n^{-1}\sum_i g_i$, and $G_S=n^{-1}\sum_{i\in S}g_i$.  If the removed correct and incorrect sets are $D_+$ and $D_-$ and $\|\nabla_\theta F_i\|\le L_i$, then
\[
\|G-G_S\|\le\frac1n\left[\sum_{i\in D_+}L_i(1-c_i)+\sum_{i\in D_-}L_i c_i\right].
\]
The first term can be small for confidently correct rows; the second makes clear that removing confident mistakes is an influence intervention rather than an approximation to their original contribution.

\subsection{Checkpoint design and scale conventions}
The FashionMNIST model is an MLP $784$--$512$--$256$--$10$ with LayerNorm. One dense-LACS seed is retrained on CPU to save checkpoints at epochs 1, 10, and 20. Eight preregistered batches of 128 examples are analyzed per checkpoint, with 20 random supports per batch for the initial random baseline. The original CE gradient is evaluated at the LACS-trained model state, not along an independently CE-trained trajectory. This distinction matters when interpreting fidelity.

For hard support normalized over survivors, $\widehat G_S=K^{-1}\sum_{i\in S}g_i$ and $G_S=(K/n)\widehat G_S$. Direction, scale, and residual are separate metrics. A near-unit cosine with a norm ratio of 1.65 is consistent with a small original-denominator residual; it is not itself small raw vector error. All corrected residuals below were calculated batch by batch, not reconstructed from mean cosines and mean norm ratios.

\begin{table}[h]
\caption{Dropped-gradient follow-up. Ranges span eight batches of one model/seed at each checkpoint; they are not confidence intervals. All early removed rows were correctly classified.}
\centering\small
\begin{tabular}{@{}crrrrr@{}}\toprule
Epoch & Mean $K/n$ & Full-$n$ rel. residual & Cosine range & Wrong drops & Cancellation factor\\\midrule
1 & .61 & .015--.036 & .9994--.9999 & 0 & .26--.33\\
10 & .61 & .004--.014 & .9999--1.0000 & 0 & .30--.36\\
20 & 1.00 & 0 & 1 & none & not applicable\\\bottomrule
\end{tabular}
\end{table}
Here cancellation factor is $\|\sum_{i\in D}g_i\|/\sum_{i\in D}\|g_i\|$. It shows that both small individual contributions and partial cancellation enter the measured residual. Continuous LACS adds a survivor-reweighting term:
\[
G_w-G=\frac1n\sum_{i\in S}(w_i-1)g_i-\frac1n\sum_{i\in D}g_i.
\]
A small dropped-gradient term alone does not bound $G_w-G$. Indeed, at epoch 20 hard support is the full set, but continuous mean-one weights still change the update substantially.

\begin{table}[h]
\caption{Initial checkpoint diagnostic, means across eight batches. Residual is relative to full-model CE; native and fitted weight scales are deliberately separate. OMP is a diagnostic on last-layer gradients, not a complete training-method reproduction.}
\centering\small
\begin{tabular}{@{}clrrrr@{}}\toprule
Epoch & Method & $K$ & Relative $L_2$ & Cosine & Norm ratio\\\midrule
1 & LACS hard, survivor mean & 77.63 & .6488 & .9997 & 1.6480\\
 & LACS continuous & 77.63 & 1.2447 & .9912 & 2.2288\\
 & Random $K$ & 77.63 & .6526 & .8370 & 1.1951\\
 & OMP fitted scale & 44.13 & .0925 & .9956 & .9826\\
 & OMP normalized weights & 44.13 & 1.8165 & .9956 & 2.8097\\\midrule
10 & LACS hard, survivor mean & 77.50 & .6560 & 1.0000 & 1.6559\\
 & LACS continuous & 77.50 & 1.7416 & .9819 & 2.7130\\
 & Random $K$ & 77.50 & .6932 & .8305 & 1.2492\\
 & OMP fitted scale & 19.75 & .2035 & .9805 & .9239\\\midrule
20 & LACS hard & 128 & 0 & 1 & 1\\
 & LACS continuous & 128 & 2.2194 & .9800 & 3.1903\\
 & OMP fitted scale & 14.13 & .2374 & .9711 & .9312\\\bottomrule
\end{tabular}
\end{table}

\subsection{What the GradMatch-style diagnostic does and does not show}
The last-layer embedding $(p-e_y)\otimes h$ with bias coordinates has 2,570 dimensions and is checked against autograd. Nonnegative ridge OMP uses regularization .5, a maximum count equal to the LACS count, and no residual-tolerance stopping. The initial implementation stops on a repeated index, so its returned count is below the cap in all 24 batches. Its smaller support is not an equal-budget comparison.

A no-repeat extension reaches the LACS count and yields support Jaccard overlap .84--.96 at epoch 1 and .69--.90 at epoch 10. With the same offline refitting of weights, mean full-model residuals are .059 for both supports at epoch 1 and .147 for both at epoch 10; equal-count random supports give .429 and .553. These are supportive representation diagnostics, not evidence that deployed LACS runs the refit solver, that OMP is inferior, or that either produces better final trained models. Normalizing fitted weights to sum one changes their scale and does not establish a GradMatch failure. At epoch 20 the learning rate is zero, making one-step validation probes vacuous; earlier probes are also norm-confounded.

\subsection{Signed and cutoff interventions}
The exact $\lambda_{\rm neg}=0$ signed variant protects all confidently wrong rows before global normalization. It normalizes the positive population separately, so it changes more than a single binary decision. Over 14 datasets, signed-minus-unsigned error is $+.1297\pm.4871$ pp (dataset SD), CI $[-.1515,.4109]$. Only four datasets favor protection. Earlier signed threshold, split-softmax, and global-softmax variants likewise did not establish a robust advantage. This supports a simpler unsigned design within these trials, not a universal statement about label information.

\begin{table}[h]
\caption{Cutoff/score study, MagicTelescope, one fixed split, three training seeds. Accuracy metric is test error mean $\pm$ seed SD; all operating points are validation-selected. Thus rows and rounds need not be equal.}
\centering\small
\begin{tabular}{@{}lrrrr@{}}\toprule
Variant & Selected setting & Test error & Active fraction & Fit rows (M)\\\midrule
Full & --- & $17.058\pm.054$ & 1.000 & 3.40\\
Native hard, max probability & $\tau=.1$ & $17.177\pm.196$ & .677 & 1.92\\
Native hard, true-label probability & $\tau=.02$ & $18.316\pm2.015$ & .682 & 2.78\\
Native hard, negative CE & $\tau=2$ & $17.058\pm.054$ & 1.000 & 3.40\\
Native continuous & $\tau=.5$ & $16.762\pm.188$ & .988 & 1.82\\
Fixed keep, max probability & keep .3 & $16.690\pm.094$ & .300 & .67\\
Fixed keep, true-label probability & keep .8 & $17.177\pm.563$ & .800 & 3.27\\
Absolute max-probability threshold & cutoff .95 & $16.809\pm.307$ & .772 & 1.33\\
Absolute true-label threshold & cutoff .8 & $17.153\pm.339$ & .591 & .97\\\bottomrule
\end{tabular}
\end{table}
The max-probability ranking's favorable comparison with true-label probability under the fixed-ratio \emph{policy} uses different selected numeric ratios (.3 and .8); it is not a matched-ratio intervention. Native negative-CE, continuous, and true-label-ratio optima reach search-grid edges. This small exploratory study establishes neither optimal hyperparameters nor a universal ranking/cutoff hierarchy.

\section{One-sided diagnostic controls: definition and evidential scope}
\label{app:onesided}
A one-sided intervention can ask whether the same unsigned high-confidence mask should be restricted to correct or incorrect predictions. Let $z_i=\ind\{\arg\max_k p(k\mid x_i)=y_i\}$, using a fixed class tie convention, and compute $q_i$ once over the full candidate population. With $d_i=\ind\{q_i\ge\kappa\}$, define the hard keep masks
\begin{equation}
 a_i^{U}=1-d_i,\qquad
 a_i^{C}=1-d_i z_i,\qquad
 a_i^{W}=1-d_i(1-z_i),\qquad
 a_i^{F}=1.
 \label{eq:onesided}
\end{equation}
Here $C$ removes only correct members of the unsigned removal set, $W$ removes only wrong members, and $U$ removes both. At a fixed model state, their removed sets form the partition $D=D_C\mathbin{\dot\cup}D_W$. The score normalization is not recomputed within either group. The $C/W$ controls are explicitly label-aware; they are not deployed versions that retain the label-agnostic guarantee.

The previous negative-side-zero study does not implement Eq.~\eqref{eq:onesided}: it applies a separate positive-side normalizer and continuous weights, followed by global coefficient renormalization. Likewise, independent model trajectories diverge after the first update, so a fixed-state partition is not an identity between masks taken from different trained models.

The new global-normalizer campaign freezes one full-data depth-1 XGBoost configuration and round endpoint per dataset, then compares Full, unsigned U, correct-only C, wrong-only W, and exact-count random controls on six datasets with three fresh splits. This is a mechanism comparison under a common base configuration, not a comparison between separately tuned algorithms. Table~\ref{tab:onesided-results} reports test error.

\begin{table}[h]
\centering\scriptsize
\caption{\textbf{One-sided global-normalizer LACS ablation.} Test error (\%, mean $\pm$ SD across three fresh splits). U removes the full high-confidence mask; C removes only currently correct members; W removes only currently wrong members.}
\label{tab:onesided-results}
\begin{tabular}{@{}lrrrr@{}}\toprule
Dataset & Full & U & C & W \\ \midrule
bank-marketing & $20.36\pm.76$ & $20.13\pm.88$ & $20.61\pm1.29$ & $20.61\pm1.11$ \\
Diabetes130US & $39.76\pm.11$ & $39.66\pm.27$ & $55.61\pm.41$ & $39.74\pm.08$ \\
electricity & $18.46\pm.28$ & $18.20\pm.21$ & $23.27\pm2.18$ & $18.23\pm.30$ \\
california & $12.05\pm.35$ & $11.93\pm.34$ & $12.09\pm.52$ & $12.07\pm.44$ \\
MagicTelescope & $16.99\pm.51$ & $16.98\pm.71$ & $16.75\pm.74$ & $17.09\pm.56$ \\
credit & $22.76\pm.71$ & $22.89\pm.89$ & $22.96\pm.88$ & $22.58\pm.42$ \\ \bottomrule
\end{tabular}
\end{table}

Using dataset as the statistical unit, U--C is $-3.583$ pp with 95\% t-CI $[-10.274,3.107]$ (permutation $p=.125$, Holm-adjusted $p=.25$), while U--W is $-.090$ pp with CI $[-.354,.173]$ (permutation $p=.469$). U--Full is $-.099$ pp with CI $[-.248,.051]$. Thus the aggregate contrasts are not statistically resolved. The diagnostic value is the pair of large C-only failures on Diabetes130US and electricity. A possible explanation is altered retained-set composition after many correct examples are removed; the experiment does not isolate that explanation from other changes along the training trajectory. Unified U removes both parts of the selected high-confidence tail and avoids those failures in these datasets. W alone is close to Full on average, so the experiment does not support a universal claim that pruning confident errors improves accuracy.

Confident errors are not merely a planted corner case in this campaign.  Averaged over rounds, currently wrong rows account for about 4.8\%, 30.3\%, 3.8\%, 1.6\%, 3.8\%, and 7.3\% of the unified removal set on bank-marketing, Diabetes130US, electricity, california, MagicTelescope, and credit, respectively; a nonempty wrong subset occurs in 92--100\% of rounds.  These are current-model mistakes, not verified semantic outliers or corrupted labels.  In contrast, the FashionMNIST same-state gradient diagnostic removes no wrong rows at the inspected checkpoints.  The mechanism is therefore data- and state-dependent.

Common-checkpoint implementation checks on planted tails and 24 reused neural batches verify the set partition and unsigned label-swap invariance. In those real neural batches, $D_W$ is empty, so $W=F$ and $C=U$; they cannot identify a benefit from removing wrong examples there.

\paragraph{Canonical churn and re-entry.}
A row-ID audit of unified LACS on Diabetes130US, california, and electricity shows that the selected support is not a permanent deletion.  Across the three trajectories, 50--67\% of rows are pruned at least once.  Among ever-pruned rows, $26.5\pm10.5\%$ re-enter on the short 13-round Diabetes run, while $71.3\pm9.6\%$ and $68.1\pm2.0\%$ re-enter on california and electricity.  Completed inactive spells have median 1.3--2 rounds and mean 1.8--9.3 rounds.  Confidently wrong rows also re-enter: among rows ever entering $D_W$, the re-entry fraction is about 22.5\%, 30.9\%, and 44.9\%, respectively.  This dynamic behavior follows from recomputing confidence; it should not be described as a formal safety guarantee.

\paragraph{Known label flips are not preferentially selected by the label-agnostic mask.}
We inject 10\% symmetric training-label corruption on Diabetes130US and electricity while keeping validation/test clean and freeze the original configuration.  The probability of entering the unified removal set is not enriched for corrupted rows: $P(D\mid\text{corrupt})/P(D\mid\text{clean})=1.05\pm.09$ and $.89\pm.09$, and confidence-decile-matched ratios are approximately one.  This is structurally consistent with label-agnostic selection.  The \emph{wrong} portion $D_W$ is different because correctness uses the observed label: corrupted examples comprise $75.5\pm1.7\%$ of $D_W$ on electricity (10\% base rate) and about 21\% on Diabetes.  Thus, when a selected row is both highly confident and wrong, label corruption can be enriched, but LACS does not directly target corruption and the experiment is not a robustness benchmark.

\paragraph{Historical versus canonical re-entry.}
An older cached long-tail selector also showed non-monotone support and about 11.5\% explicit return paths, but that implementation includes stale/random re-admission.  We therefore keep it separate from the canonical online churn statistics above.

\paragraph{InfoBatch-style boosting adaptation.}
To test whether the closest InfoBatch idea can be carried into repeated weak-learner fitting, we implement a small \emph{InfoBatch-style} adaptation rather than calling it canonical InfoBatch: rows below the current mean logistic loss are eligible, eligible rows are thinned with a fixed probability calibrated only to LACS's fit-row budget, surviving eligible fitting weights are multiplied by $1/(1-p)$, and the final 12.5\% of rounds use the full data.  The original SGD expectation argument is not transferred to nonlinear weak-learner fitting.
\begin{table}[h]
\centering\scriptsize
\caption{\textbf{InfoBatch-style boosting adaptation at a work-matched budget.} Error is mean $\pm$ SD across three frozen splits.  The adaptation obtains similar errors with a very different support; this is evidence that InfoBatch-like ideas can be ported to boosting, not that the original InfoBatch algorithm or guarantee applies unchanged.}
\label{tab:ib-boost-adapt}
\begin{tabular}{@{}lrrrr@{}}\toprule
Dataset & Full & LACS-U & IB-style & Pruned-set Jaccard U/IB \\ \midrule
Diabetes130US & $39.763\pm.105$ & $39.664\pm.269$ & $39.656\pm.233$ & $.083\pm.009$ \\
MagicTelescope & $16.987\pm.506$ & $16.975\pm.714$ & $17.307\pm.623$ & $.229\pm.005$ \\
electricity & $18.458\pm.278$ & $18.197\pm.209$ & $18.187\pm.186$ & $.231\pm.001$ \\ \bottomrule
\end{tabular}
\end{table}
At the calibrated budgets, IB-style minus LACS-U is $-.008\pm.257$, $+.332\pm.445$, and $-.010\pm.058$ pp on the three datasets.  The small pruned-set Jaccards (.08--.23) show that comparable row budgets and errors can arise from quite different supports.  Random thinning also makes re-entry far more frequent for the IB-style rule (about .72--.97) than for unified LACS (.27--.68 in the overlapping audited trajectories).

\section{Long-tail and semi-supervised boundaries}
\label{app:longtail}
\subsection{Ordinary CE and specialized losses}
\begin{table}[h]
\caption{CIFAR-100-LT ordinary-CE recipe, three seeds. The .703 figure for unsigned LACS is a \emph{nonzero-weight fraction}; this arm performs full backward and is not a measured sparse-execution gain. Other control rows are not used for speed claims.}
\centering\small\begin{tabular}{@{}lrrr@{}}\toprule
Method & Test accuracy & Nonzero fraction & Physical backward \\\midrule
Full CE & $37.95\pm0.83$ & 1.000 & full \\
Unsigned LACS weights & $38.48\pm0.38$ & 0.703 & full \\
Signed rescue & $38.83\pm0.54$ & 0.697 & not claimed \\
Random (signed partner) & $39.23\pm0.50$ & 0.697 & not claimed \\
\bottomrule
\end{tabular}

\end{table}
Unsigned-minus-CE is approximately $+.53$ pp with CI $[-.60,1.67]$, so the positive point estimate is unresolved. Signed rescue improves its point estimate further but does not outperform its own exact-count random control. That random control is partnered with the signed arm, not with unsigned LACS; it cannot isolate the unsigned support's identity effect. These are supporting supervised observations, not the central accuracy-efficiency result.

\begin{table}[h]
\caption{LDAM and LDAM-DRW combinations, mean $\pm$ SD across three seeds. The 8,000-image reporting test partition and training recipe differ from the Balanced Softmax and historical InfoBatch experiments.}
\centering\small\begin{tabular}{@{}llr@{}}\toprule
Recipe & LACS/control arm & Test accuracy \\\midrule
LDAM & dense & $40.39\pm0.09$ \\
LDAM & weight & $40.27\pm0.44$ \\
LDAM & selector & $39.77\pm0.75$ \\
LDAM & random & $39.33\pm0.62$ \\
LDAM-DRW & dense & $43.35\pm0.50$ \\
LDAM-DRW & weight & $43.67\pm0.74$ \\
LDAM-DRW & selector & $43.47\pm0.70$ \\
LDAM-DRW & random & $43.83\pm0.77$ \\
\bottomrule
\end{tabular}

\end{table}
LDAM-DRW plus LACS weighting improves by $.317$ pp, CI $[-.342,.975]$; its selector improves by $.121$ pp, CI $[-.820,1.061]$. Neither establishes an improvement. The cached selector also re-admits unseen/old rows and a random fraction, and uses a detached CE probe head rather than the specialized loss's raw logits. It is not the pure online support in Eq.~\eqref{eq:support}. Approximately .546 of candidate rows are retained, but recorded walls are confounded by concurrent jobs and are not evidence of a speedup. The probe uses a constant learning rate and some selected hyperparameters lie on a grid edge. Balanced Softmax's separate validation-frozen adaptation gives 45.57\% versus 45.87\% for the baseline and uses a fixed-ratio selector; it does not establish superiority of the normalized gate. The prior-work objectives are distinct from example selection \citep{cao2019ldam,ren2020balanced}.

\subsection{Semi-supervised development search and evaluation audit}
\label{app:ssl_audit}
The FlexMatch development setting uses CIFAR-100, WRN-28-8, 2,000 labeled training images and an intended 500-image validation carve, with 10,000 iterations under a 50,000-iteration cosine schedule \citep{zhang2021flexmatch}. The following numbers are \emph{historical recorded diagnostics under the audited loader}, not corrected full-validation or full-test benchmark estimates. Test diagnostics were computed during development; they are not used by the recorded selection rule. Earlier descriptions that test was never evaluated were inaccurate.

\paragraph{Preserved development observations.}
For canonical $\lambda=-.7,\tau=.05$, the recorded test-error differences against a count-schedule random replay are $-1.152$, $+.381$, $+2.144$, and $+3.105$ pp at 2,500, 5,000, 7,500, and 10,000 iterations. They describe one development seed and must not be interpreted as corrected multi-seed evidence. Table~\ref{tab:ssl_historical} preserves representative search values. The source-reconstructed evaluation counts below apply to the pinned loader; per-run evaluated-ID logs do not survive.

\begin{table}[h]
\caption{\textbf{Historical SSL diagnostics; validation truncation confirmed.} Values are not verified on the intended complete validation carve. Stage-1 values and the reported warm-up winner are preserved rather than silently replaced. Backward work includes labeled examples and the two unlabeled views.}
\label{tab:ssl_historical}\centering\small
\begin{tabular}{@{}rrrrrr@{}}\toprule
$\lambda$ & $\tau$ & Warm-up & Recorded val. error (\%) & Mean $K/N$ & Bwd. (M)\\\midrule
$-.4$ & .5 & 0 & 49.609 & .9967 & 9.569\\
$-.1$ & .2 & 0 & 49.609 & .9984 & 9.585\\
$-.3$ & .2 & 0 & 50.391 & .9804 & 9.423\\
$-.3$ & .1 & 0 & 51.172 & .9365 & 9.025\\
$-.7$ & .05 & 0 & 54.688 & .6567 & 6.539\\
$-.1$ & .5 & 5,000 & 48.438 & 1.0000 & 9.600\\\bottomrule
\end{tabular}
\end{table}

\paragraph{Search status and candidate ledger.}
The base grid contains 54 configurations. After the disclosed admissible-domain amendment, 42 complete in $-1<\lambda<0$, ten excluded configurations have observed empty-support initialization, and two lack failure receipts. All 24 warm-up cells subsequently complete. The original finalizer omitted the reused canonical P0 candidate and reported 65 candidates. A corrected ledger restores all $42+24=66$ and leaves the historical winner unchanged. This counting correction does not repair the truncated validation ranking.

\paragraph{Why the reported candidate cannot prune.}
The recorded accuracy and Pareto finalists coincide at $\lambda=-.1$, $\tau=.5$, warm-up 5,000. Weak-view maximum probability and the initialization score both lie in $[0,1]$. For any such vector,
\begin{equation}
 q_i=\frac{e^{c_i/.5}}{n^{-1}\sum_j e^{c_j/.5}}\le e^2<10,
 \qquad u_i=1-.1q_i\ge1-.1e^2>0.
 \label{eq:ssl_no_prune}
\end{equation}
Its hard support is therefore all rows for every admissible confidence distribution, not merely along the recorded trajectory. This proves a property of that parameter setting; it does not validate the search winner or imply a bitwise full-training trajectory when sampling order differs.

\paragraph{Evaluation truncation and required correction.}
The installed \texttt{semilearn 0.3.2} loader defaults to \texttt{drop\_last=True}. The SSL trainer requests evaluation batches of 256 without overriding that flag. Under the pinned unmodified loader this evaluates only $\lfloor500/256\rfloor256=256$ validation images and $\lfloor10000/256\rfloor256=9{,}984$ test images. The observed validation granularity $100/256=.390625$ pp is consistent with that code path. This is a source-level problem; complete remote evaluated-ID and package-identity records remain an author-side audit item, not an invented recovered set.

A new evaluation-only loader fix uses \texttt{drop\_last=False} and passes complete-coverage gates. However, no trained checkpoints or per-example predictions survive, so the full-carve ranking cannot be recovered from aggregate scores. No corrected full-validation training comparison is complete. The SSL observations therefore remain a transparent development boundary and provide no positive sparse-efficiency claim.

\paragraph{Subsequent source audit and pool correction.}
The pinned-library audit confirms the evaluation path above. Class-ordered labeled indices, a sorted carve, and an unshuffled validation loader mean the evaluated 256 images cover classes 0--50 with five images each and one image of class 51; classes 52--99 are absent. Per-run evaluated-ID logs do not survive, so this class coverage is source-reconstructed. A new trainer version sets \texttt{drop\_last=False} on evaluation loaders only and passes coverage checks on 500, 10,000, and 257 items. No trained checkpoints or per-example predictions survive from the old sweep, so its full-500 ranking cannot be repaired from summary metrics. Restoring the omitted canonical cell produces a 66-candidate ledger and leaves the historical winner unchanged, but does not repair its validation scope. A corrected rerun has not been completed. Test values were evaluated at intermediate checkpoints; the accurate claim is that test was not used for selection, not that it was never evaluated.

\subsection{Separate extensions, retained only for interpretive completeness}
An age-aware modification of InfoBatch initially gives $+.683$ pp on six CIFAR-100-LT seeds, but a four-seed replication gives $-.488$ pp. Pooled IF100 effect is $+.215$ pp with CI $[-.292,.722]$, while IF50 is negative. It is a separate temporal extension and supplies no positive evidence for the LACS confidence rule. Historical multiplicative-weighting label-noise experiments also give mixed or adverse results; they are not current hard-support robustness benchmarks. Neither branch justifies a general claim that confident mistakes are mislabeled or should always be removed.

\section{Observed selection trajectories and distribution sensitivity}
\label{app:trajectory}

\begin{figure}[h!]
\centering
\includegraphics[width=.76\linewidth]{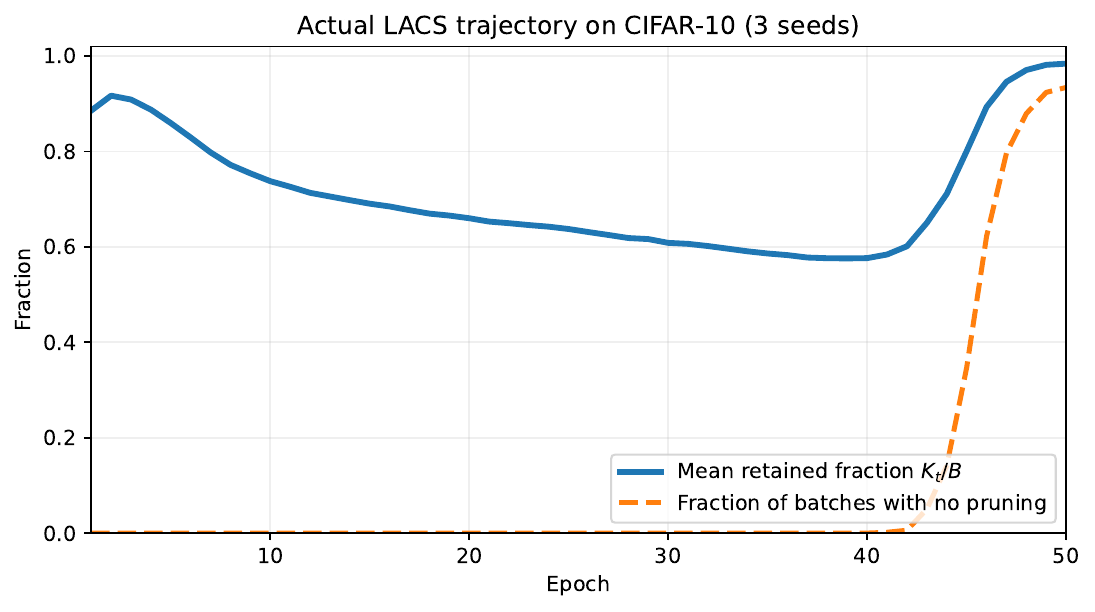}
\caption{\textbf{Actual normalized-LACS trajectory on CIFAR-10, three seeds.} The solid line is the epoch-mean retained fraction $K_t/B$; the dashed line is the fraction of batches in the epoch for which no row is pruned. The late return toward full support is predicted by the exact mass condition and is not imposed by a hand-written annealing schedule.}
\label{fig:actual-trajectory}
\end{figure}

\paragraph{Actual $K_t/n$ trajectory on CIFAR-10.}
Figure~\ref{fig:actual-trajectory} shows the measured retained fraction from the same normalized LACS rule on CIFAR-10/ResNet-32-GN across three seeds. Pruning begins in epoch 1. The epoch-mean retained fraction reaches a minimum of .573, has median .668 over the 50 epochs, and returns to .984 at epoch 50. The fraction of batches with no pruning is zero through epoch 40 and then rises to .934 at epoch 50. Thus the rule does not impose a fixed pruning schedule: support first contracts and later expands again when confidence mass accumulates near the maximum.

\paragraph{Exact empirical check of the no-pruning criterion.}
For a batch, Eq.~\eqref{eq:mass} can be checked as
\[
\frac{M_\tau(c)}{\rho n}=\frac{\exp((\operatorname{LME}_\tau(c)-c_{\max})/\tau)}{\rho}>1
\quad\Longleftrightarrow\quad D=\varnothing.
\]
Among 480 logged CIFAR-10 batches (three seeds, epochs 1/10/25/50, 40 batches per checkpoint), 108 satisfy the left-hand condition and exactly those 108 prune nothing. All 108 occur at epoch 50. The same equality holds for every logged no-prune event in two boosting trajectories: \texttt{pol} has 171 no-prune rounds and \texttt{electricity} has one, with 1.000 agreement in both cases.

\paragraph{Nearly equal means can produce different supports.}
Real batches also isolate the distribution-sensitive part of the rule. Two epoch-50 CIFAR-10 batches have mean confidences .9395 and .9411, yet their LME values are .9801 and .9824, thresholds .9979 and 1.0002, and retained fractions .625 and 1.000. Another pair has means .9404 and .9395 but retained fractions 1.000 and .625. These are not synthetic counterexamples: they are observed batches from the training trajectories. They show that the retained count is not a function of arithmetic mean confidence alone. Comparable near-mean-matched examples occur in boosting; for example, two logged rounds with means .9152 and .9194 retain .609 and .727 of their rows. These diagnostics establish distribution sensitivity of the selector; they do not by themselves establish a predictive advantage over a fixed-count policy.

\section{Reproducibility, scope, and responsible reporting}
This supplement retains the definitions, protocols, uncertainty, and controls needed to interpret the completed evidence. The supplemental archive contains a reference implementation of the normalized selector, the numerical trajectory summary used in Figure~\ref{fig:actual-trajectory}, and analytical/plotting scripts. It is a focused reproducibility package rather than a complete release of every training harness, raw benchmark artifact, or internal execution log. Unrecorded quantities, such as selected-prefix fitting time and single-run SD, remain unavailable rather than reconstructed without evidence.

Selection can disproportionately affect minority or hard examples even when it does not read labels. Accuracy aggregated over a task is therefore insufficient for deployment decisions. The long-tail class behavior, provisional semi-supervised observations, and evaluation audit motivate subgroup checks, refreshed scores, complete evaluation sets, and realistic cost accounting. The paper makes no medical or other high-stakes deployment claim. Its principal empirical conclusions concern the reported finite benchmark panels and implementations, not a guarantee that sparse selection preserves performance in new settings.

\FloatBarrier
\section{Extended related work: what is and is not new}
\label{app:related_extended}
\paragraph{Curricula, easy/hard emphasis, and robust influence.}
Self-paced learning explicitly changes which examples are presented as training progresses \citep{kumar2010selfpaced}; OHEM and Focal Loss instead emphasize hard examples and reduce the relative contribution of easy ones \citep{shrivastava2016ohem,lin2017focal}. MentorNet learns a data-driven curriculum under corrupted labels \citep{jiang2018mentornet}. These methods establish that example difficulty can be used to reshape optimization, but their direction differs from LACS on confidently wrong examples: hard-example methods retain or emphasize them. RobustBoost and ArchBoost occupy the complementary side, reducing the influence of persistently misclassified or outlying points through label-signed margins and robust nonconvex potentials/losses \citep{freund2009robust,li2018archboost}. LACS combines neither objective directly; its unsigned confidence mask can remove both confidently correct and confidently wrong rows without reading the label at the selection event.

\paragraph{Learning signals, static pruning, and dynamic reuse.}
Active Bias uses the history of true-class predictive probabilities to emphasize informative examples \citep{chang2017activebias}. Forgetting events provide another trajectory-based criterion \citep{toneva2019forgetting}; Data Diet studies early gradient-norm and prediction-error scores \citep{paul2021datadiet}. Pruning has also been studied through its effects on neural scaling, including scoring approaches that need not use class labels \citep{sorscher2022scaling}. These works preclude novelty claims for unequal example utility, label-unaware scoring alone, or improved accuracy after removal. LACS instead analyzes a particular current-population normalization and clipping rule. Dynamic pruning \citep{raju2021dynamic} and repeated random sampling \citep{okanovic2024rs2} additionally make reselection and re-entry established baselines rather than LACS contributions.

\paragraph{Loss-based pruning and recent scoring alternatives.}
Selective-Backprop uses a recent-loss percentile to set acceptance probabilities \citep{jiang2019selective}; it is already distribution-aware. InfoBatch removes a stochastic fraction of its low-loss population and compensates retained contributions \citep{qin2024infobatch}. RCAP uses class-aware allocation and probabilistic selection \citep{hassan2025rcap}. SeTa selects through loss-based difficulty groups and a moving window \citep{zhou2025seta}, while RePB addresses inconsistencies associated with score drift and training history \citep{zhou2026repb}. BLS constructs per-example importance proxies from an exponential moving average of observed batch losses \citep{zhou2026bls}; it is a scoring component that can be combined with dynamic-pruning frameworks, not evidence of a LACS comparison we have run. Against this literature, the relevant LACS question is the effect of an unsigned confidence cutoff that responds to score separation and mass. None of these relationships establishes superiority without a corresponding controlled comparison.

\FloatBarrier
\begin{table}[h!]
\caption{\textbf{How the selection boundary is determined.} ``No-prune'' describes the rule itself, not a guarantee that an implementation never samples zero rows by chance.  InfoBatch details follow its ICLR 2024 mean-loss policy; SeTa details follow the CVPR 2025 method and official implementation.}
\label{tab:selector_prior_compare}
\centering\scriptsize
\begin{tabularx}{\linewidth}{@{}p{.12\linewidth}p{.17\linewidth}p{.22\linewidth}p{.20\linewidth}X@{}}\toprule
Method & score & boundary / selection & cardinality / no-prune & retained-example treatment \\ \midrule
LACS & max predictive confidence & deterministic upper tail above log-mean-exp + margin & endogenous; no prune iff $M_\tau>\rho n$ & hard 0/1 support, or declared continuous $u_i$ variant \\ 
InfoBatch & per-sample loss & loss below arithmetic mean is eligible; each eligible row is thinned with a configured stochastic pruning probability & realized count depends on both the below-mean population and random thinning & retained low-loss rows are rescaled to correct gradient expectation; final full-data phase \\ 
SeTa & per-sample loss & configured random reduction, 1D $k$-means difficulty groups, then an easy-to-hard sliding window over groups & controlled by keep ratio, window scale, schedule, and cluster sizes; no analogous closed-form distribution condition for nontrivial settings & selected rows are used as a hard subset \\ 
Fixed quantile & chosen scalar score & order statistic at a prescribed quota & quota fixed in advance (modulo ties) & hard subset \\ 
Absolute cutoff & chosen scalar score & fixed score value & data-dependent count; no prune if all scores stay below cutoff & hard subset \\ \bottomrule
\end{tabularx}
\end{table}

\begin{figure}[h!]
\centering\includegraphics[width=.68\linewidth]{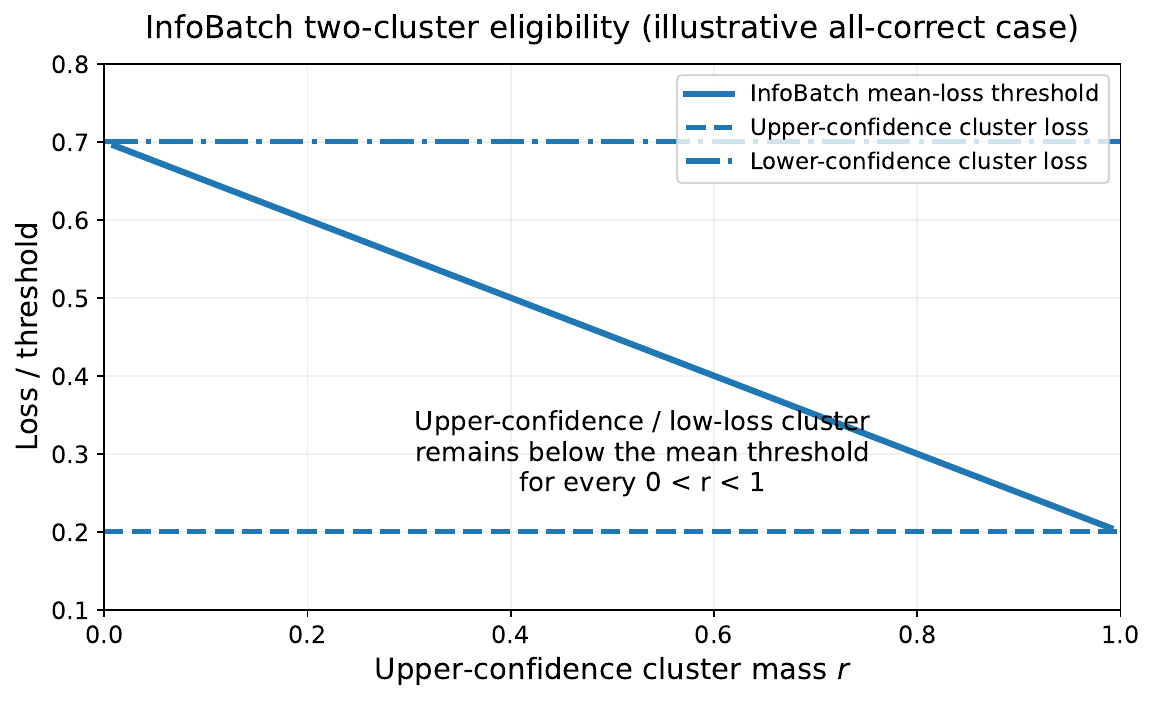}
\caption{\textbf{InfoBatch two-cluster eligibility in an illustrative all-correct case.}  If the upper-confidence cluster has lower loss than the lower-confidence cluster, the arithmetic mean loss lies strictly between the two cluster losses for every nontrivial mass $0<r<1$.  The low-loss upper cluster is therefore always eligible for InfoBatch's configured stochastic thinning; its expected removed mass is governed by the thinning probability and eligible-pool mass rather than by a LACS-style gap-and-mass no-pruning boundary.  This is a conceptual comparison of selection rules, not a claim that either rule is superior.}
\label{fig:ib-two-cluster}
\end{figure}

\paragraph{Arithmetic mean versus log-mean-exp.}
The arithmetic mean is not ``blind'' to mass: changing the fraction of high- or low-score points changes the mean.  The difference is that a mean retains only the first moment.  For the same confidence scores,
\[
\operatorname{LME}_\tau(c)=\tau\log\!\left(\frac1n\sum_i e^{c_i/\tau}\right)\ge \bar c,
\]
with equality only for a constant population.  More revealingly,
\[
T(c)-c_{\max}=\tau\log\!\frac{M_\tau(c)}{\rho n}.
\]
Thus LACS's boundary is controlled directly by the \emph{near-maximum mass}: isolated upper scores keep $M_\tau$ small and can cross the boundary, whereas a broad concentration near the maximum raises the threshold.  Two distributions can share the same arithmetic mean while having different LME because their upper-tail dispersion differs.  This is the sense in which the LACS reference is distribution-sensitive beyond a first-moment threshold; it should not be described as recovering the full distribution.

InfoBatch is therefore closer to LACS than SeTa in one important sense: both compare a current per-example score to a population-relative reference.  The references are nevertheless different.  InfoBatch's arithmetic mean partitions a \emph{loss} population into an eligible low-loss set and then prunes that set randomly; if two loss populations have the same arithmetic mean, the eligibility boundary itself is unchanged even though the composition below it can differ.  LACS uses a log-mean-exp reference plus an explicit positive margin on \emph{confidence}; even at a fixed mean confidence, its boundary can change with dispersion and upper-tail shape.  It then deterministically zeros the sufficiently separated upper tail.  InfoBatch's random rescaling is designed to recover the original gradient in expectation, whereas LACS does not impose that objective.  SeTa is distribution-dependent through loss clustering, but its operative window is determined by the configured grouping/window schedule rather than a closed-form score threshold.  Thus saying that SeTa has ``no threshold'' is best understood as \emph{no explicit scalar threshold of the LACS/InfoBatch form}; its cluster boundaries are still implicit, data-dependent boundaries.  Likewise, SeTa does not literally prune under every possible parameter setting, so the useful contrast is the absence of a distribution-derived no-pruning criterion analogous to Eq.~\eqref{eq:mass}, not an unconditional statement that it must always discard data.

\paragraph{Coordinate-aware comparison of population means.}
For a CE score evaluated at a specified model state, let $v_i=p_\theta(y_i\mid x_i)$. Then
\[
\bar\ell=-\frac1n\sum_i\log v_i=-\log\left[\left(\prod_i v_i\right)^{1/n}\right].
\]
Consequently, $\ell_i<\bar\ell$ is equivalent to $v_i$ exceeding their geometric mean. This transformed reference can change when probability dispersion changes even at a fixed arithmetic mean of $v$. InfoBatch is therefore not generally insensitive to probability-distribution shape. LACS instead applies an exponential quasi-arithmetic mean to $c_i=\max_k p_\theta(k\mid x_i)$, with an additional margin; $v_i=c_i$ only for correctly predicted rows. Actual InfoBatch scores may also be cached from different visits. The relevant distinction combines the score, its population transformation, deterministic versus stochastic removal, and survivor compensation. It is not a claim that an arithmetic loss mean uses less distributional information in every coordinate system.

\paragraph{Gradient approximation, importance sampling, and validation targets.}
CRAIG selects weighted coresets to approximate full-training gradients \citep{mirzasoleiman2020coresets}. GradMatch directly optimizes gradient discrepancy and supplies both budgeted and residual-tolerance selection formulations \citep{killamsetty2021gradmatch}. GLISTER instead uses a validation-based bilevel criterion \citep{killamsetty2021glister}. Importance sampling can use a tractable gradient-norm bound rather than full per-example gradient vectors \citep{katharopoulos2018importance}. Thus neither adaptive cardinality nor avoiding full gradient embeddings is unique to LACS. Our diagnostic asks whether the support of a different scalar rule happens to retain a useful gradient representation. Fitted versus normalized coefficients and hard versus continuous support remain distinct; a small checkpoint residual is not a full training reproduction of any of these methods.

\paragraph{Boosting and influence control.}
AdaBoost, stochastic gradient boosting, and FilterBoost established stagewise weighting or selection long before this work \citep{freund1997decision,friedman2002stochastic,bradley2007filterboost}. GOSS and MVS study sampled tree-fitting statistics \citep{ke2017lightgbm,ibragimov2019mvs}, whereas SnapBoost studies heterogeneous hypothesis classes \citep{parnell2020snapboost}. Robust boosting also makes the treatment of difficult examples a prior research question \citep{freund2009robust}. LACS does not introduce boosting by filtering, heterogeneous weak learners, or robustness through influence control. Its evaluated selection procedures use unsigned confidence with explicitly identified cutoff and fitting-weight contracts. Preserving a sampled gradient expectation does not generally reproduce a nonlinear fitting operation, and this distinction does not imply that an InfoBatch-inspired boosting adaptation is impossible.

\paragraph{The specific LogitBoost trimming precedent.}
Section 9 of \citet{friedman2000additive} trims low working weights by cumulative weight mass and permits rows to re-enter at subsequent iterations. Its LogitBoost discussion explicitly distinguishes label-free $p(1-p)$ weights from other boosting weights that depend on the signed margin. Thus the absence of $y_i$ from the selection score is not unique to LACS, and correct-only imbalance is a prior motivation as well as a useful present diagnostic. Our normalized cutoff controls exponential-confidence concentration, with the separate cardinality bound described above. Comparative accuracy requires a faithful implementation-level experiment.

\paragraph{Calibration, sparse mappings, and acquisition.}
Confidence need not estimate correctness \citep{guo2017calibration}; LACS temperature acts across examples, not class logits. Sparsemax projects scores onto the simplex to obtain sparse outputs \citep{martins2016sparsemax}; LACS instead clips weights of relatively high-confidence examples. Info-Coevolution addresses data acquisition and adaptive annotation \citep{qin2025coevolution}, rather than skipped updates on a fixed labeled set. LDAM-DRW and Balanced Softmax modify the supervised long-tail loss \citep{cao2019ldam,ren2020balanced}; FlexMatch modifies pseudo-label acceptance \citep{zhang2021flexmatch}. These objectives are related but not interchangeable with LACS selection.

\end{document}